\documentclass[pdflatex,sn-mathphys]{sn-jnl}

\jyear{2021}%

\theoremstyle{thmstyleone}%
\theoremstyle{thmstyletwo}%

\theoremstyle{thmstylethree}%

\usepackage{xcolor}

\usepackage[export]{adjustbox}

\begin{document}

\title[Democratizing Neural Machine Translation with OPUS-MT]{Democratizing Neural Machine Translation with OPUS-MT}



\author*[1]{\fnm{J\"{o}rg} \sur{Tiedemann}}
\email{jorg.tiedemann@helsinki.fi}

\author[1]{\fnm{Mikko} \sur{Aulamo}}
\email{mikko.aulamo@helsinki.fi}
\equalcont{The authors are listed in alphabetical order.}

\author[1]{\fnm{Daria} \sur{Bakshandaeva}}
\email{daria.bakshandaeva@helsinki.fi}
\equalcont{The authors are listed in alphabetical order.}

\author[1]{\fnm{Michele} \sur{Boggia}}
\email{michele.boggia@helsinki.fi}
\equalcont{The authors are listed in alphabetical order.}

\author[1,2]{\fnm{Stig-Arne} \sur{Gr\"{o}nroos}}
\email{stig-arne.gronroos@helsinki.fi}
\equalcont{The authors are listed in alphabetical order.}

\author[1]{\fnm{Tommi} \sur{Nieminen}}
\email{tommi.nieminen@helsinki.fi}
\equalcont{The authors are listed in alphabetical order.}

\author[3]{\fnm{Alessandro} \sur{Raganato}}
\email{alessandro.raganato@unimib.it}
\equalcont{The authors are listed in alphabetical order.}

\author[1]{\fnm{Yves} \sur{Scherrer}}
\email{yves.scherrer@helsinki.fi}
\equalcont{The authors are listed in alphabetical order.}

\author[1]{\fnm{Ra\'{u}l} \sur{V\'{a}zquez}}
\email{raul.vazquez@gmail.com}
\equalcont{The authors are listed in alphabetical order.}

\author[1]{\fnm{Sami} \sur{Virpioja}}
\email{sami.virpioja@helsinki.fi}
\equalcont{The authors are listed in alphabetical order.}

\affil[1]{\orgdiv{Department of Digital Humanities}, \orgname{University of Helsinki}, \orgaddress{\street{Unioninkatu 40}, \city{Helsinki}, \postcode{00014}, \country{Finland}}}

\affil[2]{\orgname{Silo.AI}, \orgaddress{\street{Fredrikinkatu 57 C}, \city{Helsinki}, \postcode{00100}, \country{Finland}}}

\affil[3]{\orgdiv{Department of Informatics, Systems and Communication}, \orgname{University of Milano-Bicocca}, \orgaddress{\street{Viale Sarca 336}, \city{Milano}, \postcode{20126}, \country{Italy}}}






\abstract{This paper presents the OPUS ecosystem with a focus on the development of open machine translation models and tools, and their integration into end-user applications, development platforms and professional workflows. We discuss our ongoing mission of increasing language coverage and translation quality, and also describe ongoing work on the development of modular translation models and speed-optimized compact solutions for real-time translation on regular desktops and small devices.}




\maketitle


\section{Introduction}\label{sec:introduction}

Language technology carries a growing responsibility in a society that is increasingly dominated by digital communication channels. Machine translation~(MT) plays a decisive role in cross-lingual information access and will continue to grow as a crucial component in our natural language processing (NLP) toolbox, enabling inclusiveness and equity among people with different cultural and linguistic backgrounds. All the major IT companies recognize the importance of MT and push significant efforts into the development of internal translation solutions with slogans like ``no language left behind"\footnote{\url{https://www.microsoft.com/en-us/research/blog/no-language-left-behind/} and \url{https://ai.facebook.com/research/no-language-left-behind/}} and similar initiatives.

\begin{figure}[tb]
    \centering
    \includegraphics[width=.8\textwidth]{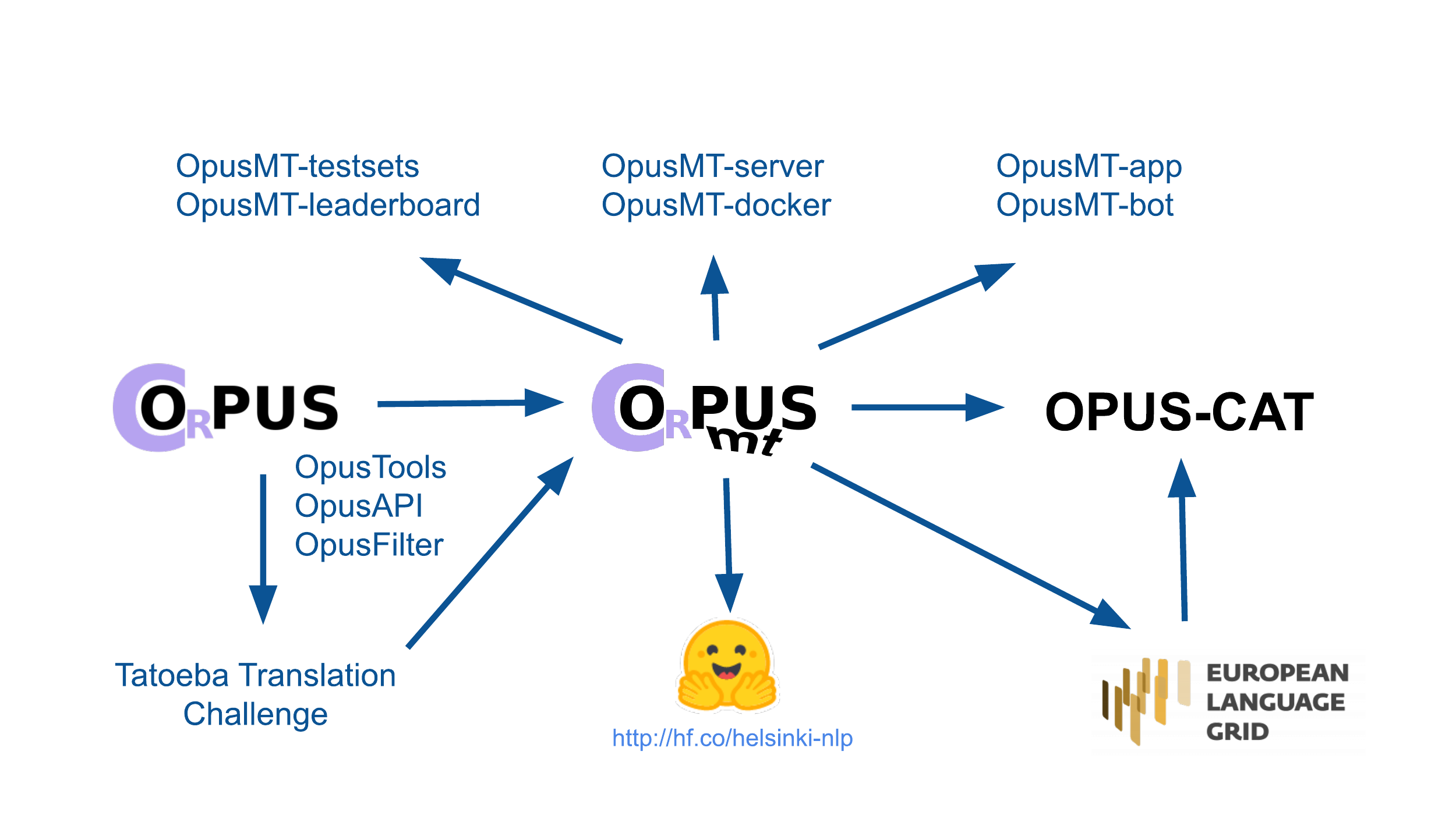}
    \caption{OPUS-MT and its connections to other components, platforms and applications.}
    \label{fig:overview}
\end{figure}

However, leaving MT to commercial exploitation comes with severe risks related to data privacy, transparency and inclusivity. The mission of OPUS~\cite{tiedemann-nygaard-2004-opus} and OPUS-MT~\cite{tiedemann-thottingal-2020-opus} is to push open and transparent solutions supported by the research community without profit-oriented goals in mind. The starting point is OPUS, a growing collection of public parallel data sets providing the essential fuel for open data-driven MT. Training data with good language coverage is crucial for high-quality MT. OPUS-MT builds on that collection and provides public translation tools and MT solutions.
The long tail of languages with limited NLP resources creates one of the largest challenges of modern language technology, and thus we aim at improved language coverage in OPUS and OPUS-MT. This paper describes the infrastructure that we are building to support our mission (see Figure~\ref{fig:overview} for a high-level overview of the various components and their connections).

Recent years have seen a revolution in natural language processing due to the advances in deep learning and computing facilities that support training complex neural network architectures. This success was only possible thanks to the availability of open source frameworks and growing public data sets. Another cornerstone in modern NLP is the distribution of pre-trained models that can be reused and adjusted to new tasks. Transfer learning has been shown to be very effective for many downstream tasks since it avoids expensive training procedures and draws significant benefits from unsupervised pre-training on raw text. Surprisingly, little has been done until recently with respect to public translation models and most teams still develop translation engines from scratch, even the most basic ones.
However, increasing energy consumption and awareness of the environmental impact of deep learning call for a better and more sustainable use of resources. We propose OPUS-MT as a major hub for pre-trained translation models along with other initiatives that distribute deployable NLP models.

The most competitive architectures for neural language and translation models also present the drawback of having an ever-increasing size. Creating and even deploying such models becomes an obstacle and can only be done by well equipped units with sufficient High Performance Computing (HPC) backbones. Thus they tend to stay in the hands of large corporations that provide MT as a service. Even when creating high-quality models efficient enough to be run locally on the end-user's device would be possible, the corporations rarely have any incentive to publish such models with a license that would permit this. 

Considering the importance of translation support, it is unfavorable if MT stays in the hands of a few high-tech corporations. Providing competitive public translation models with permissive licenses is, therefore, essential to avoid monopolies and to bring MT to the devices of end users, researchers and application developers without any strings attached and (implicit) commercial exploitation of its users. Many every-day users may not be aware of the dangers of data leaks and exploitation of personal information when using so-called free online services that feed into commercial products and targeted advertisements. 
In the spirit of open data and open source in NLP, OPUS-MT tries to democratize machine translation taking away the dependence on profit-maximizing services and tools. In the following we describe the OPUS ecosystem and how it supports this goal.

In Section~\ref{sec:opus} we focus on data outlining the principles of OPUS and the tools connected to our collection, including the OPUS-API, OpusTools and OpusFilter toolkits. The section also presents the Tatoeba translation challenge. In Section~\ref{sec:opusmt}, we provide details about the OPUS-MT framework, the training pipelines and the integration of MT models into various platforms such as Hugging Face, the European Language Grid and other end-user applications, as well as the OPUS-CAT professional workflow integration. Thereafter, we discuss benchmarks and evaluation as an important component in MT development in Section~\ref{sec:benchmarks}. Finally, we discuss our current efforts in scaling up language coverage and optimizing translation models in terms of speed and applicability in Section~\ref{sec:scaling} before summarizing related work in Section~\ref{sec:relatedwork} and concluding the paper with some final remarks.

\section{The Open Parallel Corpus OPUS}
\label{sec:opus}
OPUS\footnote{\url{https://opus.nlpl.eu}} has been a major hub for parallel corpora for about 18 years~\cite{tiedemann-nygaard-2004-opus,tiedemann-2009-news,tiedemann-2012-parallel} It serves aligned bitexts (i.e. bilingually aligned parallel texts) for a large number of languages and language pairs, providing publicly available data sets for machine translation from various domains and sources. Currently, the released data sets cover over 600 languages and additional regional language variants that are compiled into sentence-aligned bitexts for more than 40,000 language pairs. In total there are ca.\ 20 billion sentences and sentence fragments that correspond to 290 billion tokens in the entire collection. The released data sets amount to about 12 TB of compressed files. Note that each sentence can be aligned to many other languages depending on the language coverage in individual sub-corpora. The distribution over languages and language pairs is naturally skewed and follows a Zipfian curve~\cite{zipf} with a long tail of language pairs with little data, whereas only a few languages cover the main part of the data set. However, there are over 300 language pairs that contain one million sentence pairs or more, providing a good base for high quality machine translation. Note, that the collection will be skewed toward certain domains and the use of the resulting models will be affected by the domain coverage for individual language pairs as well.

\begin{figure}[tb]
Basic corpus data with sentence boundaries (abbreviated for brevity):
\vspace{.2cm}
{\footnotesize
\begin{verbatim}
<?xml version="1.0" encoding="utf-8"?>
<document>
  <CHAPTER ID="0">
    <P id="1"></P>
    <SPEAKER ID="1" LANGUAGE="DE" NAME="Rübig">
      <P id="2">
        <s id="1">Madam President, I saw a few boats landing at Parliament this ...</s>
        <s id="2">Not only were there language difficulties; the telephone line ...</s>
        <s id="3">I would be most obliged if the number on which the security ...</s>
      </P>
      <P id="3"></P>
    </SPEAKER>    
\end{verbatim}
}
\vspace{.3cm}
Standoff annotation for sentence alignment (with scores):
\vspace{.2cm}
{\footnotesize
\begin{verbatim}
<?xml version="1.0" encoding="utf-8"?>
<!DOCTYPE cesAlign PUBLIC "-//CES//DTD XML cesAlign//EN" "">
<cesAlign version="1.0">
 <linkGrp targType="s" toDoc="fr/2005/CES_AC71_2005_5SUMMARY.xml.gz" 
                     fromDoc="en/2005/CES_AC71_2005_5SUMMARY.xml.gz">
  <link certainty="0" xtargets=";1 2" id="SL1" />
  <link certainty="0.612088" xtargets="1;3" id="SL2" />
  <link certainty="0.173077" xtargets="2;4" id="SL3" />
...
\end{verbatim}
}
   \caption{An example of XML encoded OPUS data (corpus and sentence alignments). Sentence IDs are used to link sentences between two documents. Multiple sentences may appear in a link in which there sentence IDs are separated by spaces. Source and target language IDs are separated by a semicolon and empty links may also appear (like the first link in the example above where no source language sentence is aligned with sentences 1 and 2 in the target language).}
    \label{fig:opusdata}
\end{figure}

The native format for OPUS is a simple standalone XML format that allows to include additional markup depending on the original source. Sentences are marked with appropriate sentence boundary tags and alignments are stored as standoff annotation in XCES Align format. Figure~\ref{fig:opusdata} shows an example of the annotation. Aligning text files based on sentence IDs has the advantage of scaling to large massively multilingual corpora such as software localisation data sets or Bible translations. Any document can be aligned with many translations or adaptations without repeating the essential content. The general principle in OPUS is to align all available language pairs to cover every possible language pair that can be extracted from the parallel data. Even alternative alignments can be provided without modifying the original corpus files, and without any impact on other alignments. Standoff alignment can also easily be augmented with meta-information such as alignment probabilities or link type information. Those additional features can be used to filter data sets according to certain conditions (see Section~\ref{sec:opustools}).

For convenience, OPUS also provides common data formats that support a seamless integration into machine translation pipelines and downstream applications. Those files are automatically generated from the native XML format and released from the OPUS storage facilities. In particular, we provide plain text file corpora with aligned sentences on corresponding lines (referred as Moses-style bitexts due to their application in the Moses toolkit~\citep{koehn-etal-2007-moses}), and translation memory exchange files (TMX) that can be loaded from software mainly used by professional translators. Note that those derived files typically lose information such as additional markup, document boundaries as well as link type information and alignment scores. TMX files are also deduplicated (on the level of translation units, i.e. pairs of aligned sentences) and empty links are removed from both Moses and TMX file formats. The link counts reveal those discrepancies, which may result in significant differences for some sub-corpora in the collection.

For most corpora in OPUS, we also provide additional data files such as token frequency counts, word alignment files, extracted bilingual dictionaries and filtered phrase tables from statistical machine translation. We also provide monolingual data sets, some of which include data that goes far beyond the aligned texts. All of those released files are provided in a consistent file format and interface to make it convenient to integrate different subsets in systematic experiments and downstream applications. Note, however, that  depending on the information available from the original sources there are slight differences in structures, markup and language IDs used in the various data collections. Notwithstanding, we try to follow common standards with mainly ISO-639-1 language codes with fallback to three-letter ISO-639-2 and extensions specifying regional variants when necessary.

Certain inconsistencies (for example in language labels) and other issues are unavoidable since all the data sets contain different levels of noise caused by file conversion, sentence boundary detection and automatic sentence alignment, but most of it stems from the original sources. With each import, we implement filtering and cleaning procedures, and we continually aim to increase the quality level of the data. New releases will provide the result of those efforts as well as filtering software that can be adjusted for individual needs (see also Section~\ref{sec:opusfilter}).

Also note that OPUS so far focuses on sentence-level alignments in the tradition of sentence-level machine translation. Certainly, the trend goes towards increased context in any NLP application, MT included, due to the improved capabilities of modern architectures to cope with larger input and the ability to produce coherent output with long-distance dependencies across sentence boundaries. Moving the entire collection to paragraph- and document-level units is impossible as many data sets are made out of individual sentences and their translation correspondences out of any context. However, some resources already include structural information and document-level alignment and we plan to make such information more easily accessible to enable translation modelling on larger segments as well. However, so far, our efforts on machine translation are based on sentence-level models only as we will describe in more detail further down.

\subsection{Finding Data Sets using the OPUS-API}
\label{sec:opusapi}

The OPUS-API\footnote{\url{https://opus.nlpl.eu/opusapi/}} is an online API for searching data sets within OPUS. Data sets can be filtered by corpus names, available source and target languages, the type of preprocessing and the released version of the corpus. The API responds with an output in JSON format. The response contains information and simple statistics about the corpora, such as a download link to the necessary corpus files, size of the data files in bytes, the number of sentence pairs and tokens among other things. Figure~\ref{fig:opusapi-example} shows an example of the output.

\begin{figure}[tb]
{\footnotesize
\begin{verbatim}
{
  "corpora": [
    {
      "alignment_pairs": 29903792,
      "corpus": "OpenSubtitles",
      "documents": 38969,
      "id": 328013,
      "latest": "True",
      "preprocessing": "xml",
      "size": 297683,
      "source": "en",
      "source_tokens": 257890401,
      "target": "fi",
      "target_tokens": 161280297,
      "url": "https://object.pouta.csc.fi/OPUS-OpenSubtitles/v2018/xml/en-fi.xml.gz",
      "version": "v2018"
    },
    ...
  ]
}
\end{verbatim}
}
\caption{An example of OPUS-API's JSON output showing the first item when querying for the latest version of the English--Finnish OpenSubtitles corpus in XML format.}
\label{fig:opusapi-example}
\end{figure}

In addition, the OPUS-API provides the functionality to list all available corpora and languages in OPUS, as well as more specific queries. For example, one can look up all available languages for a given corpus, or all target languages for a source language within all of OPUS or within a given corpus. These commands are useful for finding data sets for specific language pairs.

\subsection{Fetching and Processing Parallel Data with OpusTools}
\label{sec:opustools}
The OpusTools package\footnote{\url{https://github.com/Helsinki-NLP/OpusTools}}~\cite{aulamo-etal-2020-opustools} enables easy access to the corpora and data files in OPUS. It provides a Python interface to the OPUS-API, methods for downloading and formatting the identified corpora, and some additional functionality for language detection and corpus splitting.

The main script in OpusTools is \texttt{opus\_read} whose primary function is to read files from OPUS and to output the sentence pairs in various supported formats.
Given source and target languages and a corpus name, \texttt{opus\_read} downloads the necessary files via information from the OPUS-API. It parses XCES format alignment files and reads the corresponding sentences from compressed XML files to produce the requested output in the desired formats, including Moses-style bitexts and TMX. The \texttt{opus\_read} script also provides the functionality to run basic 
data filtering. For example, one might only want to include one-to-one sentence alignment in the output. Additionally, it is possible to filter sentence pairs based on XML attributes such as sentence alignment confidence score (subject to the availability of the tags in the selected corpus). Furthermore, OpusTools contains the \texttt{opus\_langid} script, which can be used for language identification of OPUS files using pycld2\footnote{\url{https://github.com/aboSamoor/pycld2}} and langid.py.\footnote{\url{https://github.com/saffsd/langid.py}}~\cite{lui-baldwin-2012-langid} The script inserts language identification attribute tags in the XML files to make it possible to use language identification information for data filtering. In the future, we plan to systematically add metadata like automatically detected language tags to all data sets in OPUS. OpusTools will be useful for this task also for incoming data sets.
While OpusTools only supports basic data filtering and conversion methods, the OpusFilter toolbox described below offers a much wider range of options for processing the corpora.

\subsection{Cleaning and Preparing Data Sets with OpusFilter}
\label{sec:opusfilter}

OpusFilter\footnote{\url{https://github.com/Helsinki-NLP/OpusFilter}}~\cite{aulamo-etal-2020-opusfilter} is a
toolbox for filtering and combining parallel corpora.
The toolbox supports three main types of operations: corpus preparation tasks, essential preprocessing steps and, finally, various types of filters that aim to remove noise and unwanted content.
In contrast to tools that implement a single filtering method (e.g.~\cite{sanchez-cartagena-etal-2018-prompsit,xu-koehn-2017-zipporah}), OpusFilter allows to apply different approaches, including custom filters defined by the user.
Moreover, it can be used for other corpus manipulation tasks needed when building MT models, for example joining corpora, splitting them into training and test sets, and applying tokenization.

OpusFilter can connect to the OPUS corpus collection via the
OpusTools library, but can also operate on local text files to process any monolingual, bilingual, or multiparallel corpora.
A YAML~\footnote{\url{https://yaml.org/}} configuration file defines the pipeline to transform raw corpus files to clean training and test set files. The same pipeline can be generalized over multiple language pairs.
A single, easy-to-share configuration file design is central in our efforts to provide transparent solutions and to alleviate the reproducibility issues of experiments that affect the current MT research~\cite{marie-etal-2021-scientific}. 

\subsubsection{Corpus-level data preparation}

The first steps in parallel corpus preparation are typically related
to obtaining, joining and splitting the different corpora. The
interface to the OPUS collection via OpusTools allows automatic downloading of the existing corpora.
OpusFilter supports many common text file operations on parallel
files: concatenation, head ($N$ first lines), tail ($N$ last lines), and
slice (lines from $N$ to $M$). There is a command for taking a random
subset of a specific size from a corpus and splitting data into two subsets with
given proportion. Another command allows combining multiple translations for the same
segments in separate files into a single parallel set containing all available combinations
or a sampled subset of them.\footnote{For example, three Spanish translations and three French translations for the same English document can be combined to have nine Spanish--French pairs per each segment in the document.}
Once corpus files have been created, one can proceed with essential text preprocessing described below.

\subsubsection{Segment-level preprocessing}

The preprocessing step can be used to apply a number of segment-level
preprocessors to monolingual or parallel corpora. The implemented
preprocessors include whitespace normalization, custom regular
expression substitutions,
monolingual sentence splitting,\footnote{\url{https://github.com/mediacloud/sentence-splitter}}
tokenization,\footnote{\url{https://github.com/mingruimingrui/fast-mosestokenizer}}
Chinese word segmentation,\footnote{\url{https://github.com/fxsjy/jieba}}
Japanese word segmentation,\footnote{\url{https://github.com/SamuraiT/mecab-python3}} and
subword segmentation with BPE\footnote{\url{https://github.com/rsennrich/subword-nmt}}
\cite{sennrich-etal-2016-neural} or Morfessor\footnote{\url{https://github.com/aalto-speech/morfessor}}
\cite{virpioja-etal-2013-morfessor}.
Custom preprocessors can be added by implementing a simple Python class for them.

\subsubsection{Filtering}

OpusFilter includes two types of filters. In the scope of a corpus,
duplicate filtering provides ways to either remove identical segments
(optionally ignoring casing and non-letter characters),
either in all or just some of the languages,
within a corpus. It is also possible to remove those segments from one
corpus that overlap with another corpus (e.g.\ training data segments
that overlap with the test data).

The rest of the filters work segment-by-segment, either accepting or
removing each independently of the others. The filters currently implemented
by OpusFilter include various length-based filters,
language
identification filters (based on \texttt{langid.py}\footnote{\url{https://github.com/saffsd/langid.py}}~\cite{lui-baldwin-2012-langid}, \texttt{pycld2},\footnote{\url{https://github.com/aboSamoor/pycld2}} and \texttt{fasttext}\footnote{\url{https://fasttext.cc/}}~\cite{joulin-etal-2016-fasttext,joulin-etal-2017-bag}),
script, special character, and similarity filters,
filters using varigram language
models\footnote{\url{https://github.com/vsiivola/variKN}}
\cite{siivola-etal-2007-growing}, filters using word alignment
models\footnote{\url{https://github.com/robertostling/eflomal}}
\cite{ostling-tiedemann-2016-efficient}, and sentence 
embedding filter that estimates sentence similarity based on multilingual LASER
embeddings\footnote{\url{https://github.com/yannvgn/laserembeddings}}~\cite{artetxe-schwenk-2019-margin}.

The segment-by-segment filters that can be used in OpusFilter specify
two methods: the \texttt{score} method determines a numerical or
boolean output for the segments, and the \texttt{accept} method makes
a decision whether to accept or reject the segments with the given
score. The score may be just one value or a vector of multiple
values. For example, \texttt{LengthFilter} returns the number of
characters or words in the segment: one for monolingual corpus, two
for bilingual corpus, and so on. The minimum and maximum length
parameters determine whether the segment is accepted or not.

Instead of directly applying the filtering decisions, OpusFilter can
output the scores from the filters. They can be used to inspect the
data and determine reasonable thresholds for filtering. For this,
OpusFilter includes tools for analyzing and visualizing the
scores. In addition to direct filtering, the filter scores can also be used
to train a classifier to make the filtering decision based on a
combination of the scores~\cite{vazquez-etal-2019-university}.

\subsection{The Tatoeba Translation Challenge}
\label{sec:tatoeba}

Recently, we created a special compilation of data released under the label of the Tatoeba Translation Challenge,\footnote{\url{https://github.com/Helsinki-NLP/Tatoeba-Challenge}} (TTC for short)~\cite{tiedemann-2020-tatoeba} to overcome some of the complications present in models trained on OPUS data. OPUS is a dynamic data collection with occasional new dataset releases, hence there are plenty of ways for compiling data sets from the collection and the coverage may always change. Moreover, data preprocessing and cleaning greatly influences the quality of MT, and frequent duplicates across all subsets together with substantial noise may have a negative influence on learning procedures. 

The TTC is named after the selected test and development data, which we take from the Tatoeba corpus, a dataset of user-contributed translations in hundreds of languages. The latest release of the TTC includes 29 billion translation units\footnote{Typically, translation units refer to aligned sentence pairs but they also include units of multiple sentences aligned to corresponding translations as it is often the case in various parallel data sets. Tatoeba also frequently includes units with more than one sentence.} in 3,708 bitexts covering 557 languages. We compile the training data from OPUS corpora and focus on creating a data set with consistent language labels that can easily be used for systematic machine translation experiments. All bitexts are deduplicated and filtered. We provide randomly shuffled training sets and dedicated development and test data to support a consistent experimental setup. 

\begin{figure}[tb]
    \centering
    \includegraphics[width=\textwidth]{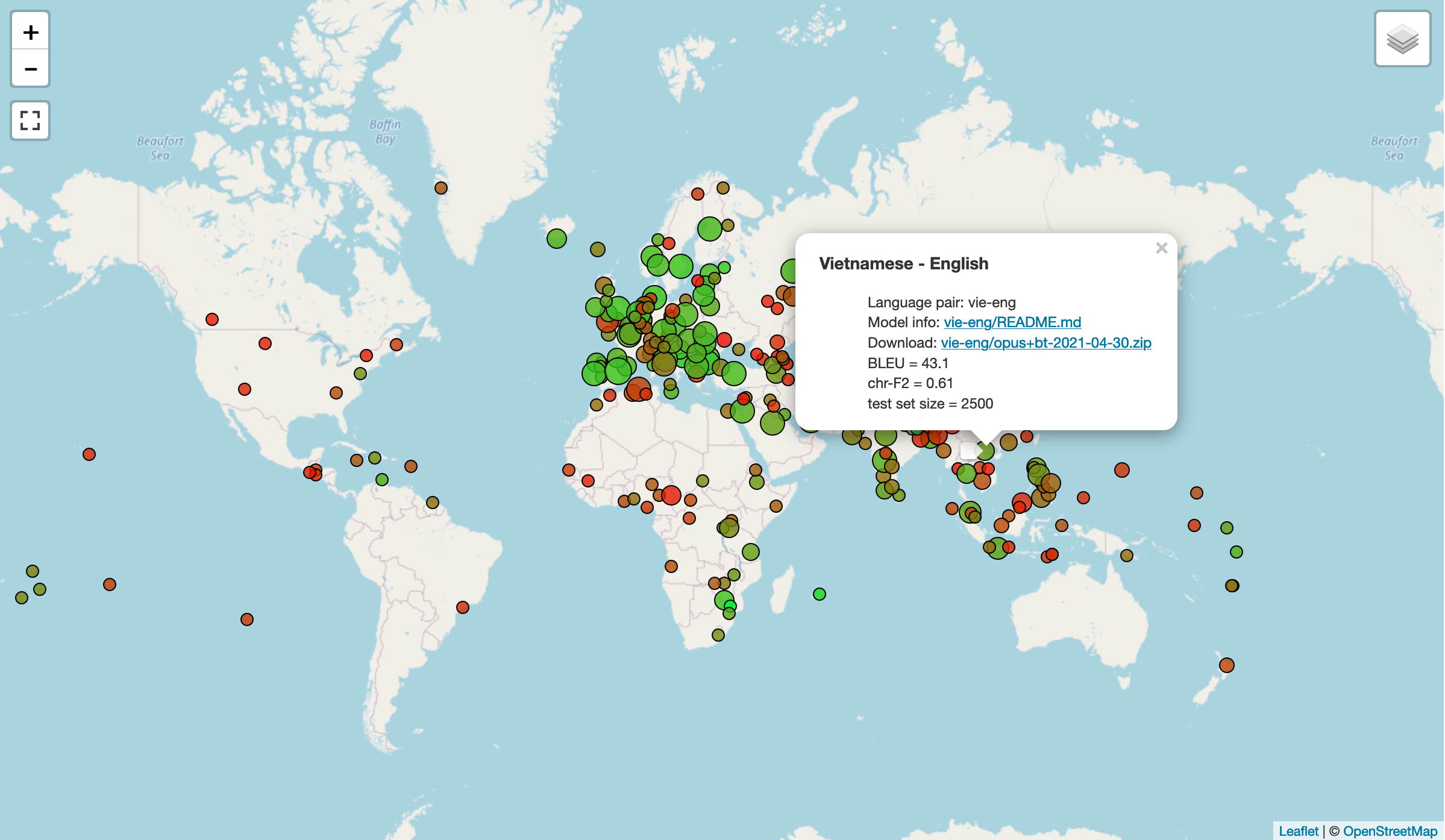}
    \caption{Language coverage of translation models visualized on an interactive map. Geolocations of languages are taken from Glottolog and dot colors indicate the translation quality in terms of an automatic evaluation metric measured on the Tatoeba test set in this case on a scale from green (best) to red (worst). Smaller circles refer to smaller, less reliable test sets.}
    \label{fig:opusmt-map}
\end{figure}

We divide the data set into several sub-challenges depending on the availability of training data. Details of the setup are available in~\cite{tiedemann-2020-tatoeba}. The current state of our machine translation development with respect to the TTC is monitored using our dashboard (see Section~\ref{sec:opusmt-leaderboard}). Geographic visualizations (see Figure~\ref{fig:opusmt-map}) also help to see the gaps in language coverage, which we try to fill in OPUS-MT. We also complement the evaluation with results from other established benchmarks. This is necessary as user contributed data such as the TTC benchmark can have various issues and a systematic quality control is beyond our capabilities. Furthermore, TTC refers to a specific kind of content mostly targeted at language learners and travelers with a focus on every-day language and rather short and frequent expressions.

We aim at regularly updating the TTC dataset in order to capture the growing support for language pairs and domains. Releases of benchmarks will be updated more frequently, depending on the growth of the original Tatoeba database. We encourage the reader to contribute to the collection of translations\footnote{\url{https://tatoeba.org/en/users/register}} to support our project. Training data releases will be compiled as needed whenever large new collections enter the OPUS collection.

\section{Open Machine Translation with OPUS-MT}
\label{sec:opusmt}

With OPUS-MT, we strive to provide public state-of-the-art translation solutions and  to be a major hub for pre-trained translation models. OPUS-MT is based on Marian, an efficient implementation of neural machine translation (NMT) in pure C++ and with minimal dependencies~\cite{junczys-dowmunt-etal-2018-marian}. Marian is a production-ready framework and includes optimized routines that enable a scalable approach to development and exploitation of modern MT systems.
In more detail, OPUS-MT targets two main objectives:

\begin{enumerate}
\item \textit{Training pipelines and models (Section~\ref{sec:training-pipelines}):} Scripts and procedures that can be used to systematically train neural MT models from public data collected in OPUS. The recipes include all necessary components that are required to create competitive translation models, including data preparation, model training, back-translation and other kinds of data augmentation, as well as fine-tuning and evaluation. Using the pipelines, models are trained on a large scale and released with permissive licenses to be reused and integrated in various platforms and workflows.
\item \textit{MT servers and integration (Sections~\ref{sec:opusmt-integration} and \ref{sec:opuscat}):} Pre-trained models need to be integrated into various platforms to make them widely applicable for end-users, translation professionals, system developers and general MT researchers. OPUS-MT provides simple server applications that can be used to deploy translation engines. The project also emphasizes integration into external infrastructures such as the Hugging Face \texttt{transformers} library and model hub (Section~\ref{subsec:huggingface}) and the European Language Grid (Section~\ref{subsec:elg}). Professional workflows are supported through plugins and installation packages tailored towards end users.
\end{enumerate}

The following sections provide an overview of the implementation and functionality of the training pipelines, the release procedures and documentation of pre-trained translation models, the implementation of server applications, and the integration into various platforms and software packages.

\subsection{Training Pipelines}\label{sec:training-pipelines}

The main purpose of OPUS-MT training pipelines is the integration of data preparation with flexible recipes for running experiments and massive NMT pre-training. Our intention is to create reusable procedures that we can run internally for creating the models we would like to use and release. However, we also share those recipes since we want to create a transparent approach that can be inspected and replicated.

The implementation of training pipelines is based on GNU Make\footnote{\url{https://www.gnu.org/software/make/}} recipes. The \texttt{make} command and makefiles are well established work horses
in the automation of compilation workflows and with that they provide a stable and tested environment, which is highly beneficial for our purposes. The philosophy of defining build targets that have various dependencies fits very well in the picture of model training and NLP experiments. Makefiles define the recipes and dependency chains that need to be followed in order to produce the final product, in our case NMT models and release packages.

There are several advantages that we benefit from when using makefiles:
\begin{itemize}
    \item Dependencies are automatically resolved and the build process is interrupted if essential dependencies cannot be resolved.
    \item Intermediate files and results can be reused and do not have to be built again if not necessary. Timestamps automatically determine whether targets need to be re-built and updated, which may affect the entire pipeline.
    \item Targets can be built in parallel and the \texttt{make} command determines which recipes can run simultaneously without breaking dependencies defined by the internal recipes.
    \item Unnecessary intermediate files are deleted at the end of the process.
    \item Static pattern rules can be used to create template recipes, which can help to define generic procedures that apply to many related tasks.
    \item Many in-built functions support the manipulation of variables, file names and build workflows.
\end{itemize}

OPUS-MT tries to make use of those advantages and implements recipe chains that cover all necessary sub-tasks for training, testing and releasing NMT models. The goal is to provide various high-level \texttt{make} commands that can run in batch mode with variables that control properties and procedures to enable a systematic exploitation of massively parallel data sets. Without those generic recipes it would not have been possible to train such a large amount of models that we were able to release already (over 2,300 models at the time of writing, 744 multilingual ones among them with varying language coverage on source and target side).

Our training pipelines are stored in a public git repository\footnote{\url{https://github.com/Helsinki-NLP/OPUS-MT-train}} and build recipes are subject to change as we continuously develop the system. The main components and principles stay the same and documentation is provided within the repository. Here, we only provide a high-level overview of the package and refer the interested reader to the original source.

\subsubsection{Setup and Basic Training}

Setting up and starting training batch jobs is straightforward and is optimized for modern Linux-based systems with sufficient storage space, the availability of a CUDA compatible GPU and access to online sources. Installing the pipelines and software dependencies can be done on the command-line using the installation recipes:

\vspace{.3cm}
{\small
\begin{verbatim}
git clone https://github.com/Helsinki-NLP/OPUS-MT-train.git
cd OPUS-MT-train
git submodule update --init --recursive --remote
make install
\end{verbatim}
}
\vspace{.3cm}

Training generic models can be done by basic high-level targets that take care of the entire pipeline and sub-tasks necessary for fetching data and starting the training procedures.
The documentation includes an extensive list of command line variables to control the properties and parameters of the various sub-tasks.
However, the most essential variables to be set are the source and target language specifications:

\vspace{.3cm}
{\small
\begin{verbatim}
make SRCLANGS="fi et" TRGLANGS="da sv en" data
make SRCLANGS="fi et" TRGLANGS="da sv en" train
make SRCLANGS="fi et" TRGLANGS="da sv en" eval
\end{verbatim}
}
\vspace{.3cm}

The above example demonstrates how easy it is to create a (multilingual) neural OPUS-MT model by just giving a set of language IDs to be included in the model (Finnish and Estonian as source languages and Danish, Swedish and English as target languages in this case). Note that the first command can be skipped because \texttt{data} is a prerequisite for the training recipe that will fetch and prepare data in any case.\footnote{Note that the \texttt{eval} target is not implemented with \texttt{train} as a pre-requisite in order to avoid that calls for evaluation automatically trigger expensive data fetching and training procedures.} The recipes are designed to take care of fetching all available data sets from OPUS with combinations of source and target languages. A random disjoint sample (in all language combinations) from a dedicated corpus will be used as validation and test data and remaining data goes into training. There is a manually specified hard-coded priority list of OPUS corpora that will be used as test and development data focusing on language coverage and relative cleanliness. The list and mechanisms are likely to change over time with further evolvements of OPUS and the OPUS-MT training recipes. Alternatively, the TTC data sets can be used to apply a consistent split into train, development and test data. Models that use multiple target languages will automatically use target language labels to augment the completely shared transformer model. 

In general, OPUS-MT will use the latest releases of OPUS to prepare the data.
Some basic sanity checking and filtering (removal of non-printable or broken Unicode characters and some length-based filtering) is done but no major changes are applied to the sources. Subword segmentation models are trained using SentencePiece~\cite{kudo-richardson-2018-sentencepiece} and word alignments are created using \texttt{eflomal}~\cite{ostling-tiedemann-2016-efficient} to feed the guided alignment feature that is used by default.\footnote{Neural word aligners like awesome aligner \citep{dou2021word} could be used as well but we prefer a language-agnostic aligner like \texttt{eflomal} that does not require pre-trained language models with a certain language coverage and the potential need of additional fine-tuning.} The default parameters provide a good fit for regular baseline models but we encourage further optimization of hyper-parameters. For more details on how to control the setup, please look at the documentation of the recipes themselves. The \texttt{eval} target will compute automatic evaluation metrics on heldout data. Another target (\texttt{eval-testsets}) can be used to also benchmark with other test sets available for supported language pairs. Those need to be available in the repository.

OPUS-MT also supports temperature-based data sampling~\cite{arxiv.1907.05019}, which is important to balance the availability of specific language pair examples in multilingual translation models. Adding a parameter to the training data creation recipes will influence the proportions used for skewed data sets. To give a practical example: adding \texttt{MAX\_DATA\_SIZE=1000000 DATA\_SAMPLING\_WEIGHT=0.2} to the \texttt{make} command will cause the system to select proportionally to the temperature-adjusted likelihood $p_l^{1/5}$ of observing an example in language $l$ from the entire data set. The additional parameter \texttt{MAX\_DATA\_SIZE} sets the sample size to one million examples for the largest language pair and all other language pairs will be down- or over-sampled according to the sample rate.

\subsubsection{Batch Jobs on HPC Infrastructure}

The main mode of running OPUS-MT training pipelines is to use batch jobs on high-performance computing (HPC) clusters using the Slurm Workload Manager\footnote{\url{https://slurm.schedmd.com/overview.html}} as a job scheduler. To support a flexible use of all recipes, OPUS-MT implements generic pattern rules to turn any recipe (or makefile target) into a Slurm batch job. This is done by defining a special suffix that can be appended to any target available in the makefile-based build system. When adding such a suffix, the recipe will generate a Slurm script with the original target as a goal and submit it to the job scheduler using the setup specified for the HPC-specific environment. Note that those settings are now basically hard-coded to work in our own environment and adjustments need to be made to match the local installation of your HPC cluster. Below, you can see an example for submitting the training job for the same model discussed above with a walltime of 72 hours (the \texttt{.submit} suffix triggers the creation of the batch job script and the submission to the Slurm engine):

\vspace{.3cm}
{\small
\begin{verbatim}
make SRCLANGS="fi et" TRGLANGS="da sv en" WALLTIME=72 train.submit
\end{verbatim}
}
\vspace{.3cm}

Various variables can be used to control the behavior and resource allocation of those jobs. For details, have a look at the documentation and definitions in the source code.

\subsubsection{Data Augmentation}

Data augmentation is particularly important for less resourced languages. However, even better resourced languages benefit from back-translation~\cite{sennrich-etal-2016-improving} and other techniques that increase the coverage of the data. OPUS-MT implements convenient procedures to produce back-translations and pivot-based triangulation data.

\begin{figure}[tb]
    \centering
    \includegraphics[width=\textwidth]{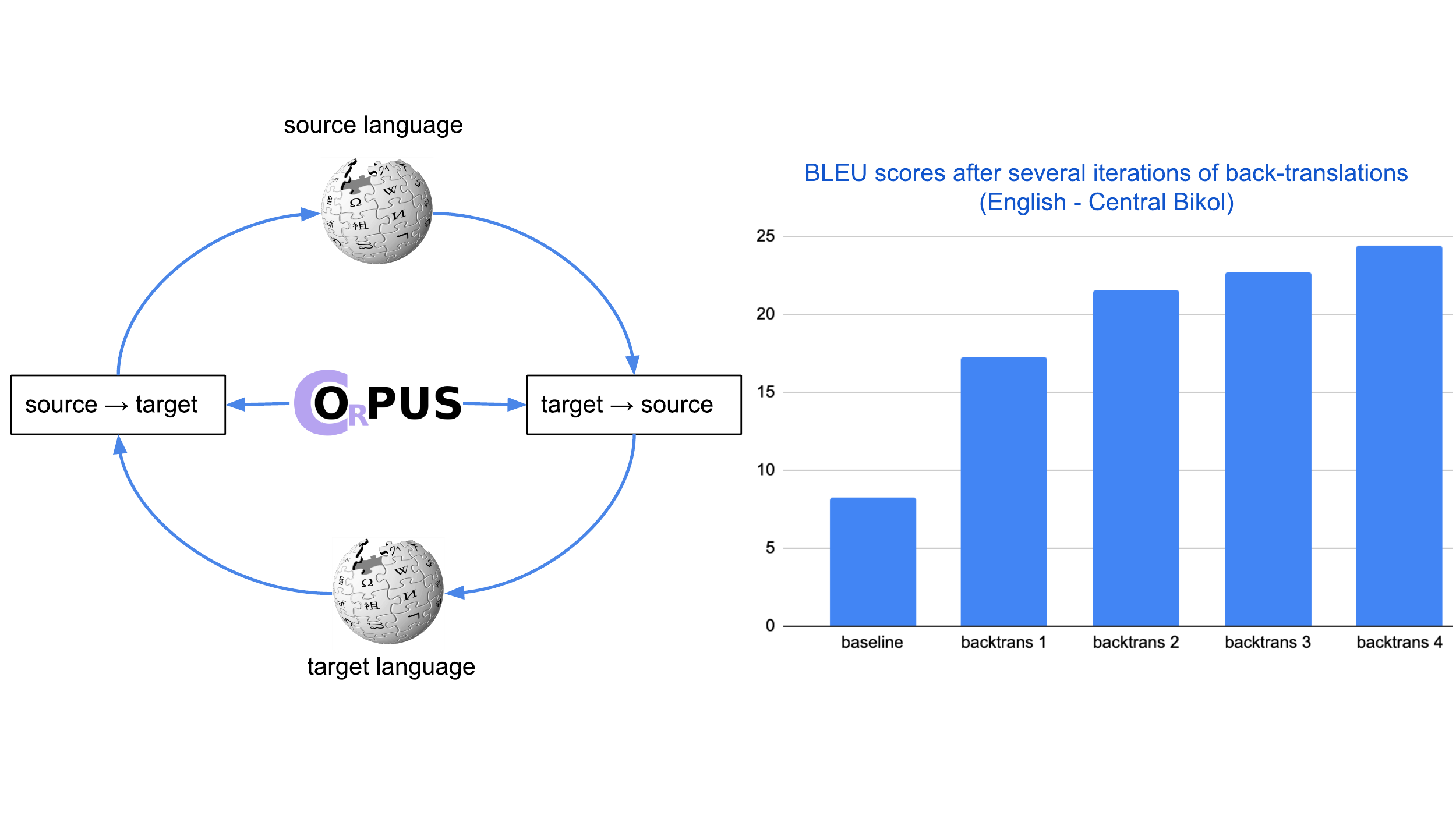}
    \caption{Iterative back-translation as a means of data augmentation. OPUS-MT uses Wikipedia content as monolingual data and the example illustrates the iterative improvements (in the barchart to the right) for translations between English and Central Bikol (an Austronesian language spoken in the Philippines) in terms of BLEU scores. Monolingual data coming from Wikipedia is translated in several rounds of improved forward and backward translation using the fresh translations to create the current synthetic back-translated data for augmenting the translation models in both directions.}
    \label{fig:opusmt-backtrans}
\end{figure}

By default, monolingual data extracted from Wikipedia and other public data sets provided by the Wikimedia foundation are available for back-translation. For supported languages, OPUS-MT fetches the previously prepared data from our object storage and uses existing models in the opposite translation direction to translate them back into a source language. Back-translation is done for individual language pairs, but multilingual models can certainly be used besides of bilingual ones for that purpose, too.

Training OPUS-MT models in different directions can be done iteratively and, thus, back-translated data can improve step by step and with that the quality of translation models in both directions can go up \citep{hoang-etal-2018-iterative}. This is especially useful in low-resource settings where initial back-translations may be quite noisy and translation quality first needs to reach reasonable levels in both directions. An example of such an iterative back-translation procedure is illustrated in Figure~\ref{fig:opusmt-backtrans}, where a model for translating from English to Central Bikol is improved with several iterations of back-translation.

\begin{figure}[tb]
    \centering
    \includegraphics[width=\textwidth]{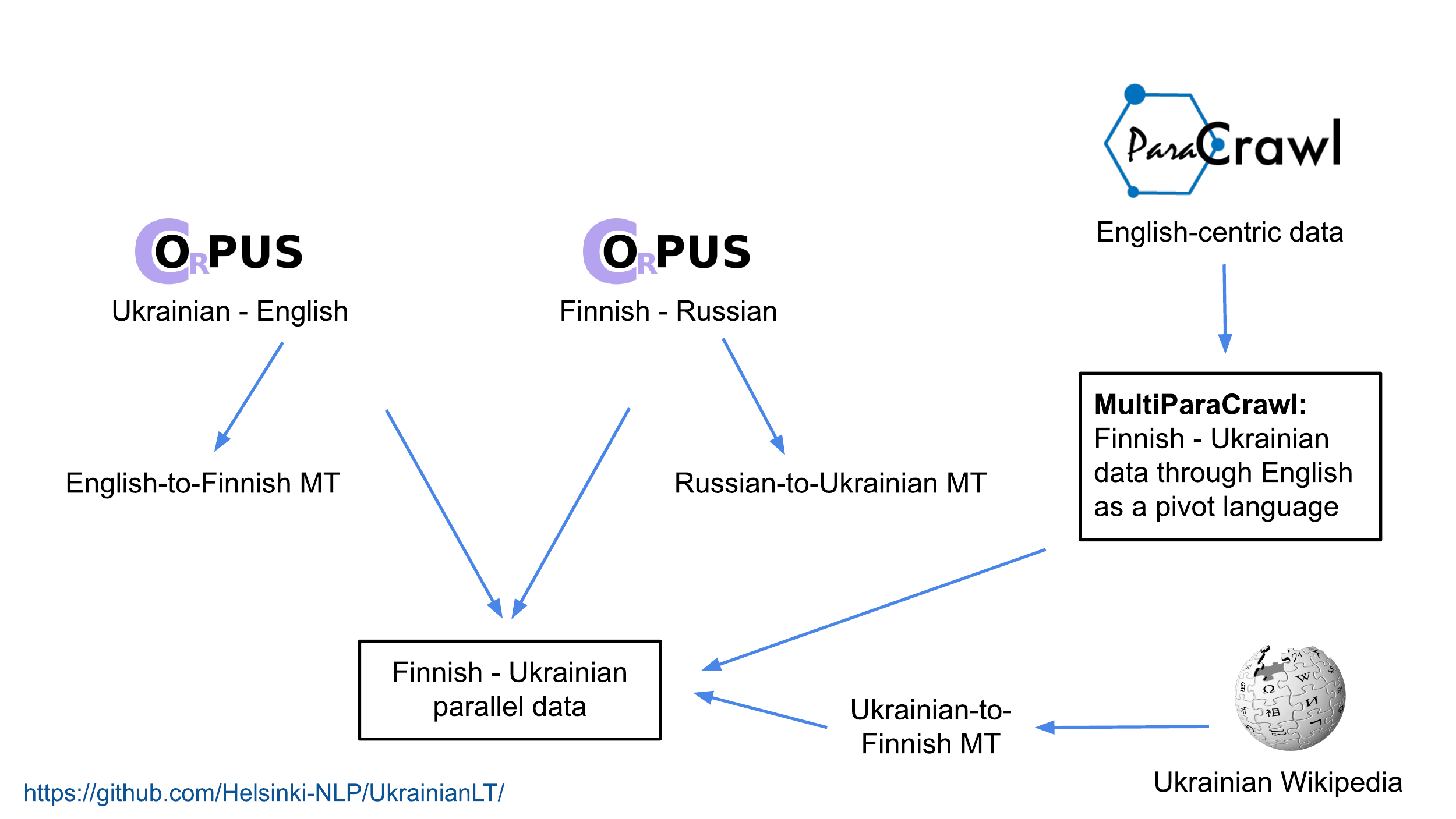}
    \caption{Data augmentation using triangulation and pivot-based machine translation. Synthetic training data can be created from English-centric parallel corpora and automatic translations of some pivot language using existing OPUS-MT models. The illustration shows the approach on the example of Ukrainian--Finnish translation.}
    \label{fig:opusmt-ukrainian}
\end{figure}

Another effective method for data augmentation is triangulation and pivot-based translation. Many data sets are English-centric, but the demand for direct translation between non-English languages is certainly growing and still under-explored. A practical real-world example arose in the on-going crisis in Ukraine. Refugees moving to various European countries need support to manage information flow and communication in the local languages. Figure~\ref{fig:opusmt-ukrainian} illustrates the various ways of augmenting data for improved translation, in this example between Ukrainian and Finnish (in both directions).

Triangulation of English-centric data is an easy way of producing additional training material. MultiParaCrawl\footnote{\url{https://opus.nlpl.eu/MultiParaCrawl.php}} is created in this way by pivoting alignments on identical sentences in English. A non-negligible amount of the 1.5 million sentence pairs for Finnish--Ukrainian could be extracted in this way from the original English--Ukrainian and English--Finnish bitexts in ParaCrawl.

Pivot-based translation is another straightforward way of producing artificial training data. Translating one side of a bitext can turn existing data sets into a synthetic parallel corpus for a different language pair. This is especially useful if strong high-quality models can be used to perform that automatic translation. In our running example, we can use a multilingual model for related East Slavic languages to translate from Russian to Ukrainian and another optimized model for translating English into Finnish to transform English--Ukrainian and Russian--Finnish data into the desired Finnish--Ukrainian language pair. Note that we include both synthetic source language data (from the English--Ukrainian corpus where English is translated to Finnish) and synthetic target language data (from the Russian--Ukrainian corpus with Russian translated into Ukrainian). Typically, synthetic target language is not preferred because of the additional noise that may enter a translation model. However, given the quality that can be achieved for closely related languages, this procedure is stable enough and quite effective as another means of data augmentation.

Multilingual models would be an alternative for implicitly including pivot languages, but the pivot-based direct translation approach has the advantage that we do not need to increase the capacity of the translation model (by blowing up the number of parameters) to accommodate additional languages nor do we need to be concerned about balancing between different languages in mixed language data. Furthermore, we can once again benefit from existing pre-trained models that have already been optimized in various ways, avoiding to re-learn everything from scratch.

\subsubsection{Fine-Tuning}

Usually, it is also common to fine-tune models for specific tasks or domains. OPUS-MT supports fine-tuning in a way that a short cycle of additional training based on small task-specific training data can be triggered. A common situation is that a local translation memory can serve as additional training material. Special recipes enable such a fine-tuning process.

Another common scenario is that multilingual models can be fine-tuned for specific language pairs. The idea is to take advantage of transfer learning in the general training phase and then to use language-specific data to fine-tune the model for that particular language pair. This is also supported by OPUS-MT and the released collection of training pipelines and recipes. However, in our experience, it is difficult to define proper learning parameters to avoid over-fitting and catastrophic forgetting. Therefore, we typically do not release such fine-tuned models and leave it up to end-users or developers of specific solutions to perform such a process.

\subsubsection{Evaluating and Releasing}

The final task we want to discuss in this section refers to evaluation and packaging of models. The targets for testing the models have already been mentioned earlier. The OPUS-MT repository, furthermore, includes common benchmark test sets and those are useful to compare model quality with other established and published research. We are currently working on a more principled way to benchmark all released models. More information about that effort is given in Section~\ref{sec:benchmarks}.

An important feature is also the creation of release packages. Special targets are defined to collect all necessary information, generate documentation in terms of readme files and YAML files together with benchmark results and other essential details. Release packages contain models including essential pre- and post-processing scripts, subword segmentation models, and decoder configurations. Training log files are also included. For internal use, we also support convenient functions for storing and fetching models as well as intermediate work files from our external storage service. Various procedures are defined to update information available for external users including lists of released models and important properties that define their functionality.

Releases are done inside of ObjectStorage containers with a CEPH backend,\footnote{Ceph is an open-source, distributed storage system, see \url{https://ceph.io/}.} which are linked from project websites and GitHub markdown pages. Information about their performance is also available and more information on benchmarking will be given further down in Section~\ref{sec:benchmarks}.

\subsection{Machine Translation Server Applications}

OPUS-MT also provides ways of deploying models and building translation servers to demonstrate and use MT from web applications or through service APIs. Our implementations\footnote{\url{https://github.com/Helsinki-NLP/OPUS-MT}} leverage the efficient decoder released as part of Marian-NMT\footnote{\url{https://marian-nmt.github.io/}} in its server mode to create a running translation service. Two alternative applications are available, one based on the Tornado web framework and another one implementing websocket services that can easily be deployed on common Linux servers and virtual machines.

The Tornado-based solution is also containerized using docker and can in this way be deployed in various environments. Simple web frontends demonstrate the usage of the service APIs. Configuration and setup is straightforward using the released packages from OPUS-MT. The server backends take care of all basic pre- and post-processing (including sentence splitting and subword tokenization) and provide a simple JSON interface to the actual server. Both applications can also combine several translation services and provide them through the common interface and API. Models to be supported can be specified in configuration files. An example for a Tornado web app configuration is shown in Figure~\ref{fig:opusmt-tornado}.

\begin{figure}[tb]
    \centering
    {\footnotesize
\begin{verbatim}
{
    "en": {
        "es": {
            "configuration": "./models/en-es/decoder.yml",
            "host": "localhost",
            "port": "10001"
        },
        "fi": {
            "configuration": "./models/en-fi/decoder.yml",
            "host": "localhost",
            "port": "10002"
        },
    }
}    
\end{verbatim}
}
    \caption{A simple configuration for an OPUS-MT server using the Tornado web app implementation providing services for English--Spanish and English--Finnish machine translation.}
    \label{fig:opusmt-tornado}
\end{figure}

The websocket service provides some additional functionalities that can be useful when deploying OPUS-MT models. There is an additional router daemon that can serve as a central access point, which is able to connect to various translation backends. The integration of multilingual models is also well supported. Furthermore, the output also provides the segmentation into subword units and an alignment between them, which is meaningful for models that are trained with the guided alignment feature of Marian-NMT.\footnote{This feature allows to guide one of the cross-lingual attention heads in the transformer with pre-computed token alignments. In this way, we obtain an attention pattern that specializes on word alignment and it also helps to kickstart the training procedures with relevant prior knowledge, which can be useful especially in low-resource settings.} Interfaces can then show links between input and output tokens to better trace the connections between source and target language inside of the translation model. Alignments can also be useful for additional post-processing features built on top of the translation service itself like placing tags and formatting information. An example output is shown in Figure~\ref{fig:opusmt-websocket}. Furthermore, the server backend also implements a cache that speeds up translation substantially when identical sentences are passed into the engine. The cache implements a simple yet efficient lookup function in a hashed database, which is stored on the local disk of the server.

\begin{figure}[tb]
    \centering
    {\footnotesize
\begin{verbatim}
{
    "alignment": [
        "0-0 1-1 2-2"
    ],
    "result": "Huomenta, Suomi",
    "segmentation": "spm",
    "server": "192.168.1.15:40002",
    "source": "sv",
    "source-segments": [
        "\u2581Godmorgon , \u2581Finland"
    ],
    "target": "fi",
    "target-segments": [
        "\u2581Huomenta , \u2581Suomi"
    ]
}
\end{verbatim}
}
    \caption{A translation response from a websocket server using OPUS-MT to translate from Swedish to Finnish. The final result after post-processing is available in the ``result" item and ``alignment" refers to subword token alignments between source and target language segments. Source and target segments list subword tokens separated by whitespaces. The Unicode character \texttt{\textbackslash u2581} at the beginning of some of them indicates the connection to the previous token (i.e.\ the need for removing the preceding space), which is commonly used by the SentencePiece tokenizer~\citep{kudo-2018-subword}.}
    \label{fig:opusmt-websocket}
\end{figure}

Our web applications are mainly provided for demonstration purposes, but they can be used as a starting point for serious platform integration and real-world applications, as we will discuss in the next section.

\subsection{Platform Integration}\label{sec:opusmt-integration}

The web services and applications described in the previous section already demonstrate the practical use of OPUS-MT models 
beyond the pure NLP research field. But even in research, platform integration becomes increasingly important: training is expensive and developing models from scratch becomes rarer in current approaches based on transfer learning. 
Integration into popular libraries and platforms is, therefore, essential to avoid unnecessary overheads of getting started with the basic facilities needed to make progress in research and development.

In this section, we describe the efforts of connecting OPUS-MT to various external platforms and software packages in order to make our models widely available and accessible for end-users, application developers and basic NLP researchers. We only provide a few examples of possible use cases. Many other applications are potential platforms where OPUS-MT can be integrated. For example, our models are already available through Tiyaro.ai,\footnote{\url{https://console.tiyaro.ai/explore?q=opus-mt&pub=Helsinki-NLP}} another model hub for AI apps. The resources are also listed in Meta-Share\footnote{\url{https://metashare.csc.fi/}} and tools could also be added to the CLARIN switchboard.\footnote{\url{https://switchboard.clarin.eu/}}

\subsubsection{PyTorch and the Transformers Library}
\label{subsec:huggingface}
The success of deep learning has been made possible thanks to the availability of open software that allows easy adaptation of neural approaches to a wide range of tasks, NLP-related ones being very visible among them. General-purpose frameworks such as PyTorch\footnote{\url{https://pytorch.org/}} and Tensorflow\footnote{\url{https://www.tensorflow.org/}} provide the essential backbone of most of the work done in this area. Specialized high-level libraries on top of those frameworks nowadays make it easy to get started with state-of-the-art approaches to neural NLP and also enable access to released pre-trained models of various kinds. The \texttt{transformers} library published by Hugging Face\footnote{\url{https://huggingface.co/transformers/}} is one of the most popular hubs of modern NLP, which is largely driven by the community.

\begin{figure}[tb]
    \centering
    \includegraphics[width=.8\textwidth]{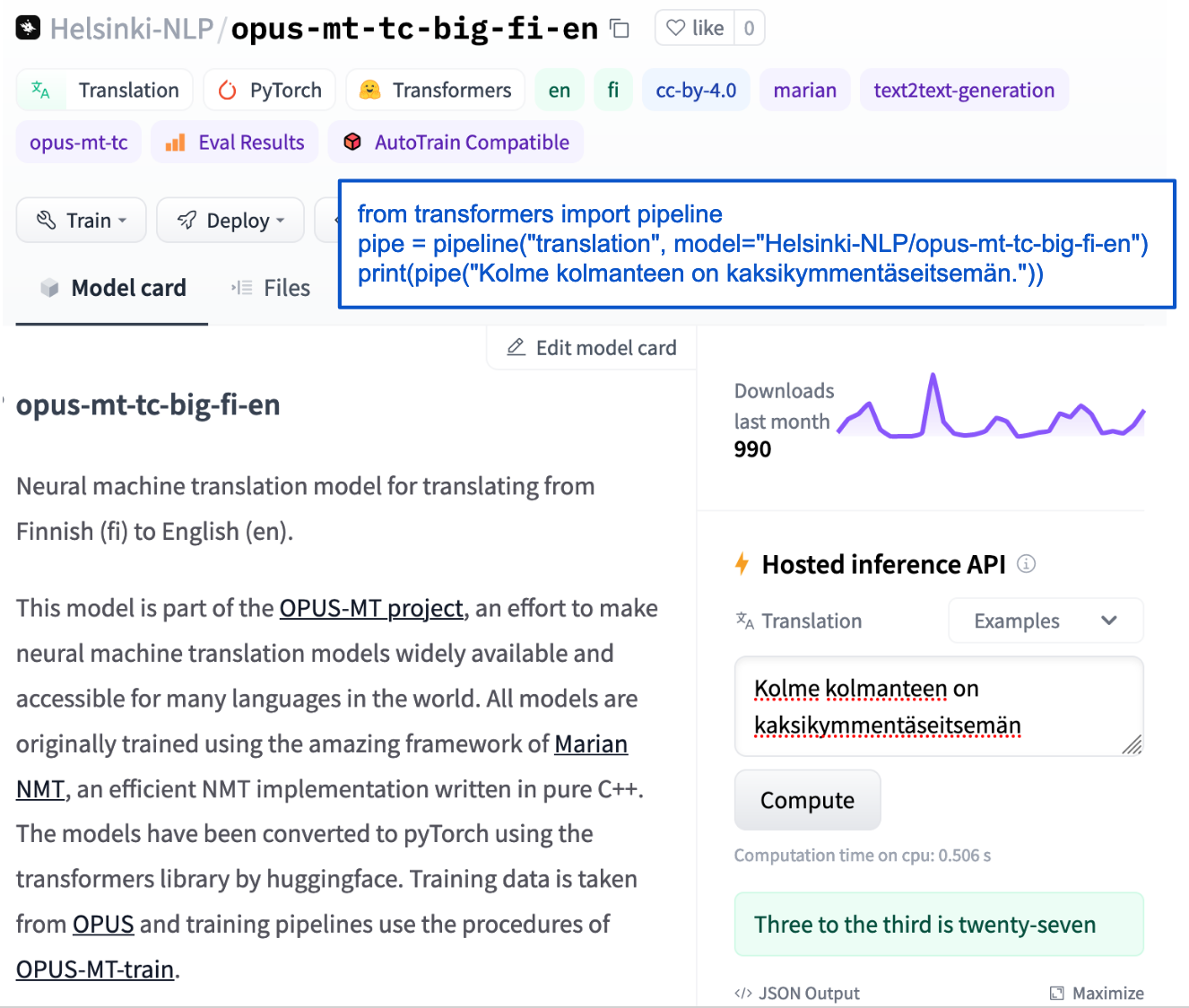}
    \caption{An example of an OPUS-MT model card in the Hugging Face model hub. The blue text box on top of the screenshot shows a three-line code snippet for using the model from the \texttt{transformers} library. The model card includes information about the use of the model, supported languages and links to relevant project websites with further information about the original model and data sets used for creating it. The model hub also provides a live inference API that can be used to test the model (in the right column) and download statistics are also shown on the top of that column.}
    \label{fig:opusmt-hf}
\end{figure}

With the help of the scientists at Hugging Face, OPUS-MT models have been fully integrated into the \texttt{transformers} library by converting them to PyTorch. The impact is significant as this enables a wide range of users to immediately get access to thousands of pre-trained translation models supporting many languages and language pairs. Models are now available from the Hugging Face model hub and can be used with a few lines of code or even through the online inference API (see Figure~\ref{fig:opusmt-hf}).

The collaboration with Hugging Face is on-going and future developments will make their way into the popular framework. Recently, conversion tools were adapted to cover more flexibly different architectures. Tensorflow-based backends are also supported now, which creates additional opportunities.

\subsubsection{Integration into the European Language Grid}
\label{subsec:elg}

The European Commission has been one of the most important players in creating resources and solutions for inclusive language technology. The European Language Grid (ELG) is one of the EU-supported initiatives to build a comprehensive infrastructure for NLP resources and tools.\footnote{\url{https://live.european-language-grid.eu/}} OPUS-MT has been funded as one of the ELG pilot projects. This has led to a seamless integration of translation services based on OPUS-MT models in containerized server implementations at ELG.

ELG services can be accessed from their live platform and models are loaded on demand from their cloud infrastructure running through Kubernetes\footnote{\url{https://kubernetes.io/}} and OpenStack.\footnote{\url{https://www.openstack.org/}} HTTPS requests can be sent to the internal API and services such as OPUS-MT can also be called programmatically from, for example, a dedicated ELG python library. Metadata records and persistent identifiers based on Digital Object Identifiers (DOI)\footnote{\url{https://www.doi.org/}} create sustainable resources according to the Findability, Accessibility, Interoperability, and Reuse (FAIR) principles.\footnote{\url{https://www.go-fair.org/fair-principles/}} The collaboration with ELG ensures long-term preservation of our developments and provides the necessary maintenance through their standardized infrastructure.

OPUS-MT now includes procedures to generate metadata records and docker images that can be pushed directly to Docker Hub\footnote{Docker Hub is a repository of user-contributed software container images, see \url{https://hub.docker.com/}.} and ELG. Through those routines, new services can easily be registered inside of the ELG platform and become available to end-users and developers after some internal validation period. Docker images are naturally also available outside of the ELG platform and can be fetched and deployed locally or on other cloud services. We also integrate ELG translation services into OPUS-CAT, making it possible to include OPUS-MT in professional translation workflows. More information on OPUS-CAT can be found in Section~\ref{sec:opuscat}.

\subsubsection{End-user Applications}
\label{sec:applications}

The previous sections already showed several ways of integrating OPUS-MT into end-user applications through online services and containerized server solutions. Further integration into tools described below demonstrate additional use cases and application areas.

\begin{figure}[tb]
    \centering
    \includegraphics[width=.8\textwidth]{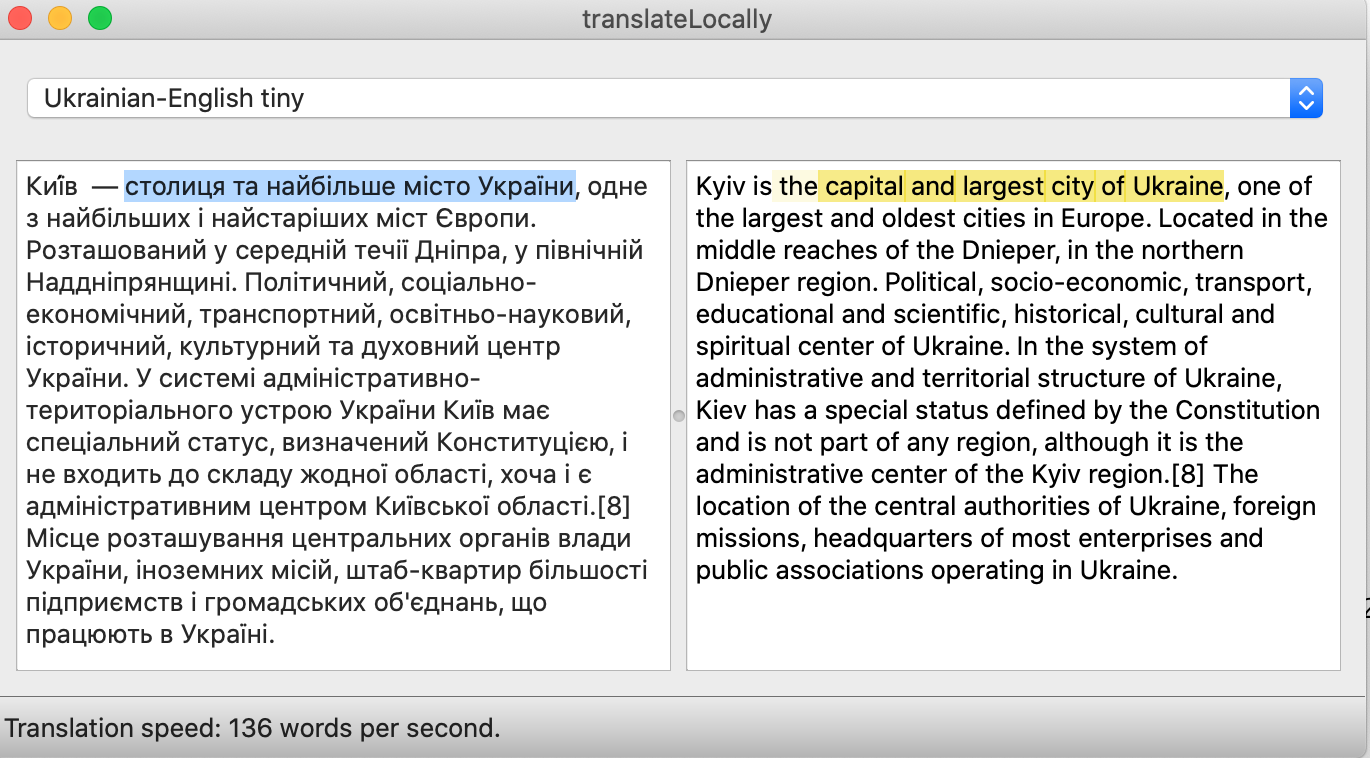}
    \caption{The translateLocally desktop application developed in the Bergamot project with an adaptation for OPUS-MT models. Here showing the example of Ukrainian--English translation with some cross-lingual highlighting done through the alignment feature.}
    \label{fig:opusmt-app}
\end{figure}

\textbf{Interactive and instant translation} is useful for quick access to information in other languages. The Bergamot project\footnote{\url{https://browser.mt/}} created speed-optimised implementations for local translation engines that can run inside of a web browser or in dedicated desktop applications.\footnote{\url{https://translatelocally.com/}} The main idea is to use knowledge distillation, quantization and lexical shortlists\footnote{A lexical shortlist restricts the output vocabulary to a small subset of translation candidates to improve decoding speed, see \url{https://marian-nmt.github.io/docs/} for further information.} to push the limits of decoding speed. Furthermore, decoder implementations can be optimized in various ways and a customized fork of Marian creates the backbone of the Bergamot solution. OPUS-MT models can be used through the same infrastructure as they are natively built in Marian-NMT. Furthermore, we currently work on systematic distillation of OPUS-MT models to create efficient student models that are compatible with the Bergamot project (see also Section~\ref{sec:opusmt-distill}). With those, our models become available in the repository of their browser-based MT solutions\footnote{\url{https://translatelocally.com/web/}} and the translateLocally desktop app\footnote{OPUS-MT fork at \url{https://github.com/Helsinki-NLP/OPUS-MT-app/}} (see Figure~\ref{fig:opusmt-app}).

\begin{figure}[tb]
    \centering
    \includegraphics[width=.3\textwidth]{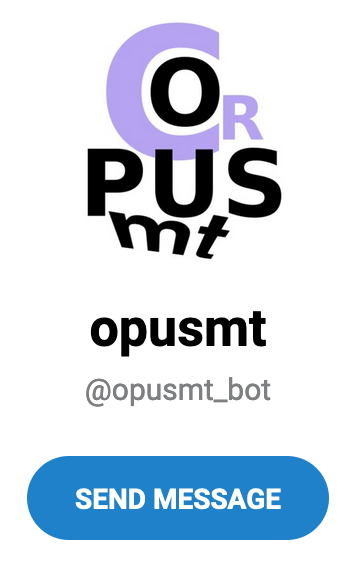}
    \includegraphics[width=.5\textwidth]{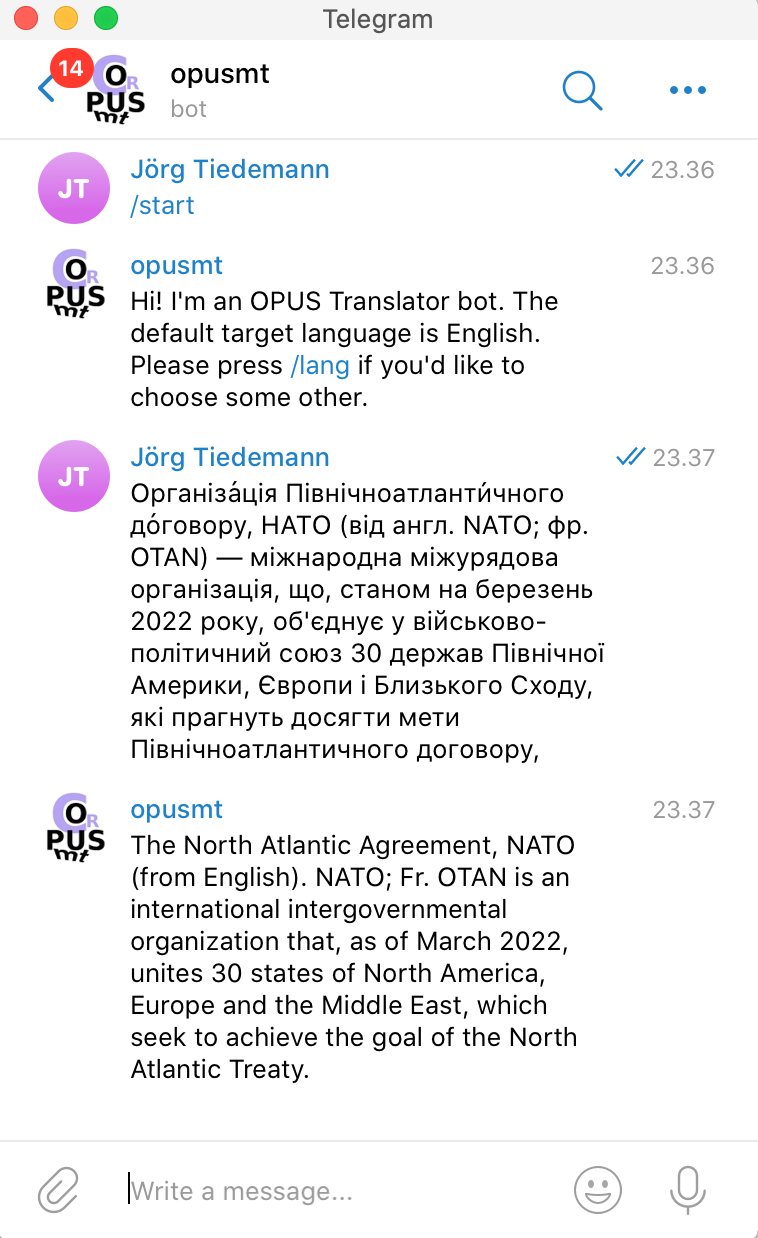}
    \caption{A simple Telegram translation bot serving the translation from and to Ukrainian. The example shows the translation of a short text snippet taken from Ukrainian Wikipedia.}
    \label{fig:opusmt-telegram}
\end{figure}

\textbf{Plugins and add-ons} for commonly used applications are another means of bringing OPUS-MT to the actual end users. One example is social media channels that are frequently used by millions and even billions of people around the World. As a response to the crisis in Ukraine, we developed a prototype for translating from and to Ukrainian using a translation bot in Telegram. To date, according to \mbox{SimilarWeb} statistics\footnote{\url{https://similarweb.com/apps/top}} Telegram has been the most-used messaging platform in Ukraine both in Google Play Store and Apple App Store. The bot is implemented using \texttt{aoigram}, a framework for Telegram Bot API written in Python; it is asynchronous, thus multiple requests can be processed almost simultaneously. Interacting with the bot is easy and convenient as sending a message is all that is required to obtain the desired translation. The bot uses our websocket server and supports several source and target languages in connection with Ukrainian. An example of the operation of the bot is shown in Figure~\ref{fig:opusmt-telegram}.

\subsection{Professional Workflows with OPUS-CAT}
\label{sec:opuscat}
OPUS-CAT\footnote{\url{https://helsinki-nlp.github.io/OPUS-CAT/}} is a collection of software packages that enable translators to use pre-trained OPUS-MT models in computer-assisted translation tools. MT is currently routinely used in professional translation work, but the field is dominated by proprietary MT systems offered by large tech companies (such as Google or Microsoft) or specialized machine translation vendors (such as DeepL and ModernMT). OPUS-CAT offers a free and open-source alternative to the proprietary MT products.

OPUS-CAT consists of a local MT engine and a selection of plugins and other types of CAT tool integration. The core of the local MT engine is a Windows build of the Marian framework, which is supplemented by a GUI for downloading and automatically installing OPUS-MT models from a centralized repository. The engine can be simply installed by extracting the installation package, and models can be searched by language names. The local MT engine does not require a connection to any external service, all the translations are generated on the user's computer using its native CPU. The local MT engine is currently available only for Windows, since many CAT tools are also only available in Windows, and professional translators are therefore likely to be Windows users. Linux or Mac versions of the MT engine are not planned at the moment, as they would probably not expand the potential user base of OPUS-CAT significantly.

The MT engine GUI provides a simple functionality for translating text, but the translations are mainly intended to be generated via an API that the MT engine exposes. CAT tool plugins and other integrations can be built on top of this API. OPUS-CAT currently supports most of the widely-used CAT tools. It includes dedicated plugins for three desktop CAT tools: Trados Studio\footnote{\url{https://www.trados.com/products/trados-studio/}}, memoQ\footnote{\url{https://www.memoq.com/}}, and OmegaT\footnote{\url{https://omegat.org/}}. In some CAT tools, such as Wordfast\footnote{\url{https://www.wordfast.com/}}, OPUS-CAT can be used by connecting directly to its API through a custom MT provider functionality. OPUS-CAT also includes a Chrome browser extension, which makes it possible to use OPUS-MT in browser-based CAT tools. The Chrome extension currently supports Phrase\footnote{\url{https://phrase.com/}} (formerly Memsource) and XTM\footnote{\url{https://xtm.cloud/}}, and modifying it to support other browser-based CAT tools is relatively simple.

One of the advantages of using OPUS-CAT in professional translation is that it is inherently secure and confidential. Since no external services are used, sensitive data is never at risk. While many commercial MT providers offer on-premises installations similar to OPUS-CAT, such installations are expensive and they cannot be adapted as freely as the open-source OPUS-CAT. The guaranteed confidentiality of OPUS-CAT also makes it possible for individual translators to use it in their work without breaching confidentiality agreements.

OPUS-CAT is intended for professional translators, so it includes functionalities for addressing issues related to MT use in professional translation, such as domain adaptation and tag handling.

The utility of generic NMT models in professional translation is uncertain, while performance improvements resulting from the use of domain-adapted NMT models have been observed multiple times \citep{laubli-etal-2019-post,informatics7020012}. Because of this, OPUS-CAT MT Engine includes a mode for fine-tuning models with bilingual data. Since fine-tuning has to be performed locally on a CPU, only small amounts of data are used (usually tens of thousands of translations pairs) and the training only lasts for a single epoch. To avoid problems with over-fitting, a very conservative learning rate is used (0.00002, a fifth of the default initial learning rate in Marian).

During fine-tuning, the model is validated against in-domain and out-of-domain validation sets, and the validation results are displayed graphically for the user to allow them to detect potential problems. Despite the conservative training settings and the small amount of data, informal testing and user feedback indicates that the fine-tuning has a noticeable effect on the usability of the MT system.

In professional translation, source documents often contain placeholder tags or tag pairs indicating formatting. Placing tags manually is time-consuming, so MT systems designed for professional translation should ideally place source tags automatically in the generated MT. OPUS-CAT supports two methods of placing tags automatically. When fine-tuning a model, OPUS-CAT can be specified to include tags in the fine-tuning set, enabling the fine-tuned model to learn the correct tag placement implicitly. With base models, sub-word alignments (supported by most OPUS-MT models) can also be used to deduct the correct placement of tags.

OPUS-CAT further supports rules that can be used to pre-edit the source text before using it as MT input, or to post-edit the MT output before presenting it to the translator. These rules can be applied to address systematic errors in source texts or in MT output. For instance, a pre-edit rule can be created to change the letter case of the source text or to correct recurring typos, and a post-edit rule can be used to correct a recurring MT mistake.

Deviation from client- or domain-specific terminology is one of the main obstacles to using MT in professional translation. While fine-tuning and the edit rules help to address some of the issues, they have their limitations: fine-tuning does not guarantee the use of correct terminology, it only increases the tendency to use it, and edit rules can be feasibly defined for only a very limited amount of cases. Work is currently in progress to include stronger terminology support in OPUS-CAT by implementing soft terminology constraints based on target lemma annotations \citep{bergmanis-pinnis-2021-facilitating}.

OPUS-CAT is currently the only open-source, free solution for neural machine translation use in professional translation. There is clearly demand for such a solution, as many individual translators currently use OPUS-CAT in their work, and several organizations have included OPUS-CAT in their translation workflows. From the point of view of the wider OPUS-MT project, the significance of OPUS-CAT is that it provides another channel of disseminating the pre-trained models and of gathering feedback and experiences from an important group of MT users, i.e.\ professional translators.

\section{Benchmarks and Evaluation}
\label{sec:benchmarks}

Important in development but also for deployment is quality control and procedures to monitor progress. Benchmarks and evaluation pipelines are essential to fill that need. Fortunately, regular shared tasks in machine translation produce various benchmarks and evaluation strategies, and recently growing interest in low-resourced languages and domains also improves the language coverage of available test sets. Within our ecosystem, we try to systematically collect existing benchmarks and contribute to the collection also with our own efforts, e.g., through the Tatoeba translation challenge mentioned earlier in Section~\ref{sec:tatoeba}.

\begin{figure}[tb]
    \centering
    \includegraphics[width=.8\textwidth]{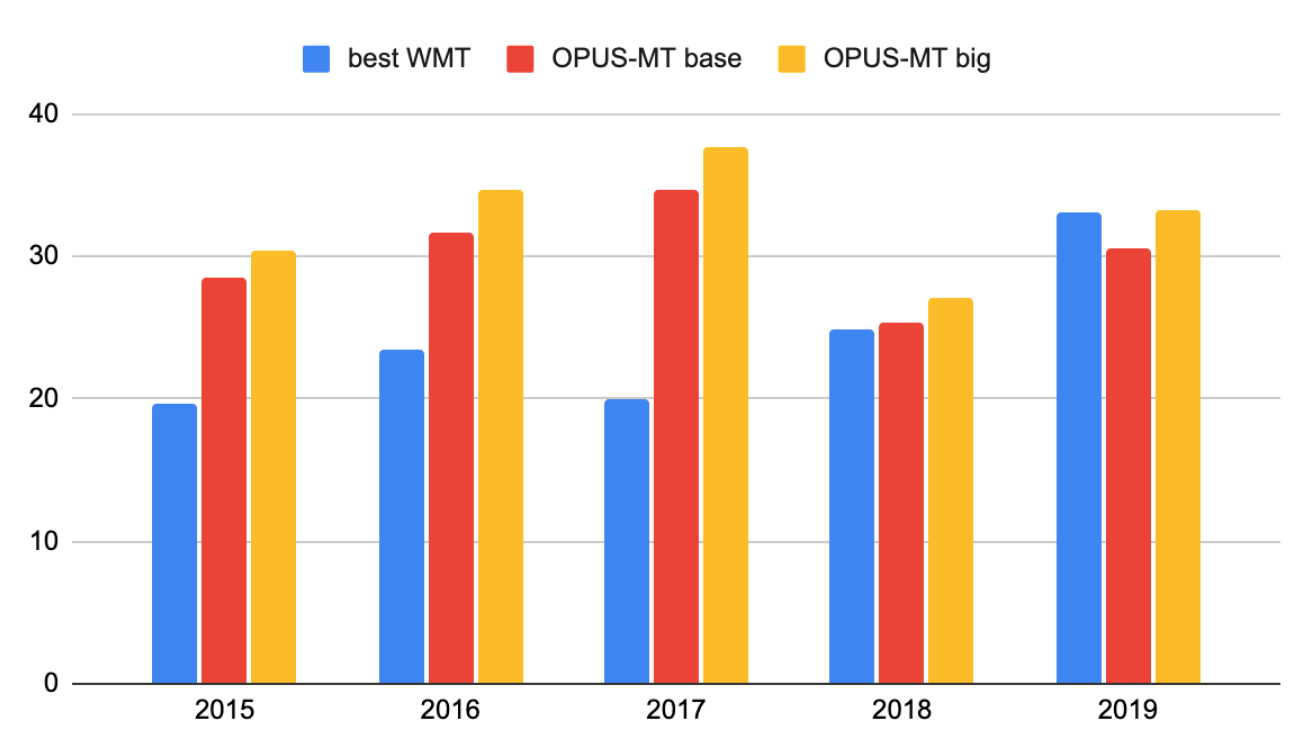}
    \caption{Comparing OPUS-MT models (in terms of BLEU scores) to official results from the news task at WMT in several years for translating from Finnish to English. WMT scores are taken from \url{http://wmt.ufal.cz/}.}
    \label{fig:opusmt-results}
\end{figure}

One of the crucial questions for the success of OPUS-MT is the quality of the models we release. Regularly monitoring and widely evaluating them is therefore essential. Comparing our models to established benchmarks and test sets is one way of approximating translation quality. OPUS-MT does not try to compete with highly domain-optimized models submitted to specific shared tasks but rather focuses on general-purpose models that can be re-used and refined later. Nevertheless, putting our models in perspective with other results is a good way of demonstrating their applicability. Figure~\ref{fig:opusmt-results} shows the example of Finnish--English translation results measured on the popular news translation task at WMT. The figure shows that our models fare well (in terms of BLEU scores) in comparison to officially submitted systems during the evaluation campaign even though they are not directly comparable for various reasons (for example, being trained on different data sets and tuned for different domains).

For practical reasons, we currently focus on automatic evaluation but we also discuss options where community-driven leaderboards can facilitate regular manual evaluations as well. To support systematic benchmarking, we compile test sets into our own repository,\footnote{\url{https://github.com/Helsinki-NLP/OPUS-MT-testsets/}} which feeds into the OPUS-MT leaderboard described further down in Section~\ref{sec:opusmt-leaderboard}. Additionally, we have also developed test suites to complement the general-purpose translation quality assessment addressed by regular MT benchmarks, described in Section~\ref{sec:test-suites}.

\subsection{The OPUS-MT Dashboard}
\label{sec:opusmt-leaderboard}

\begin{figure}[tb]
    \centering
    \includegraphics[height=6.5cm,valign=t]{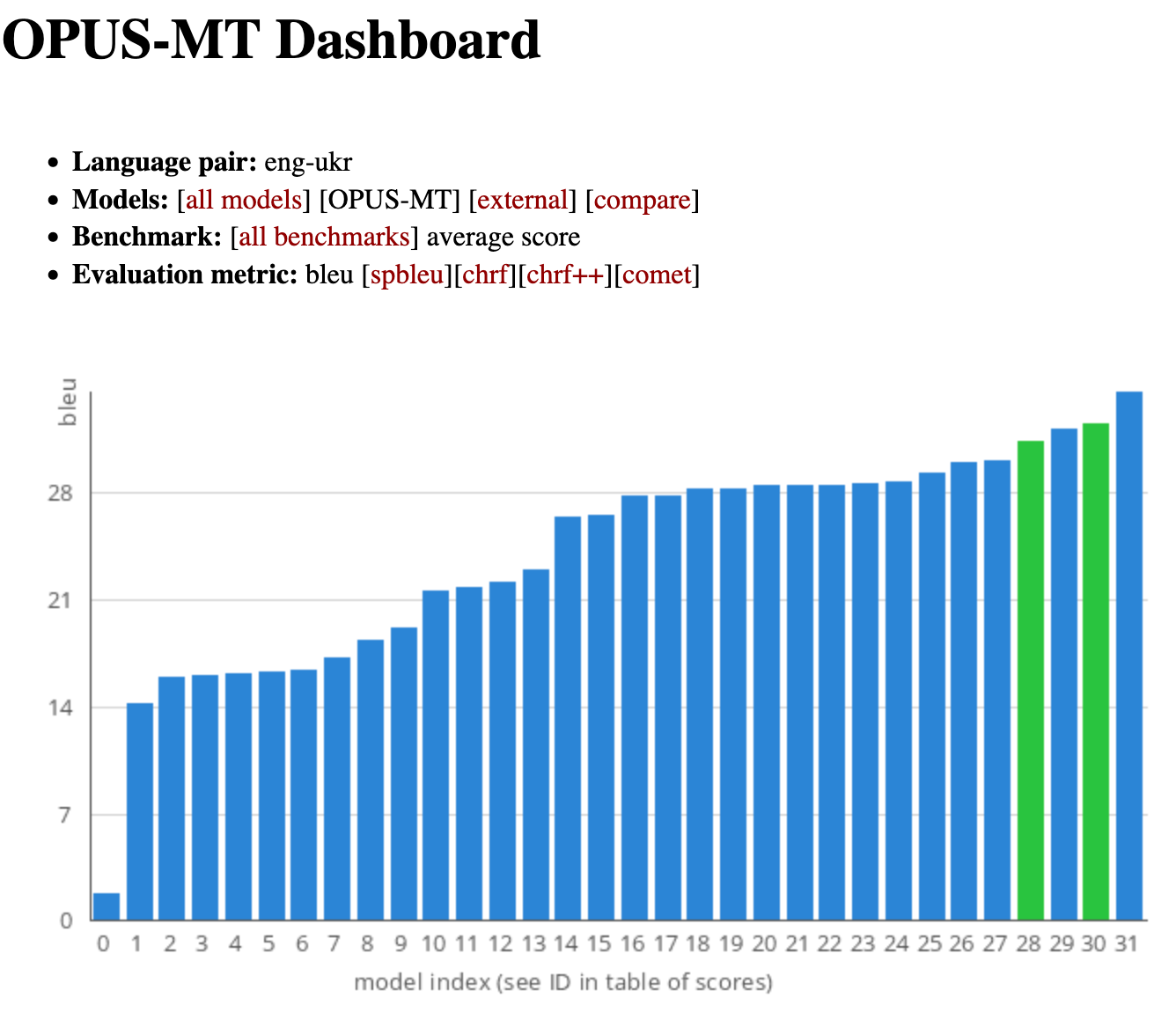}
    \includegraphics[height=6cm,valign=t]{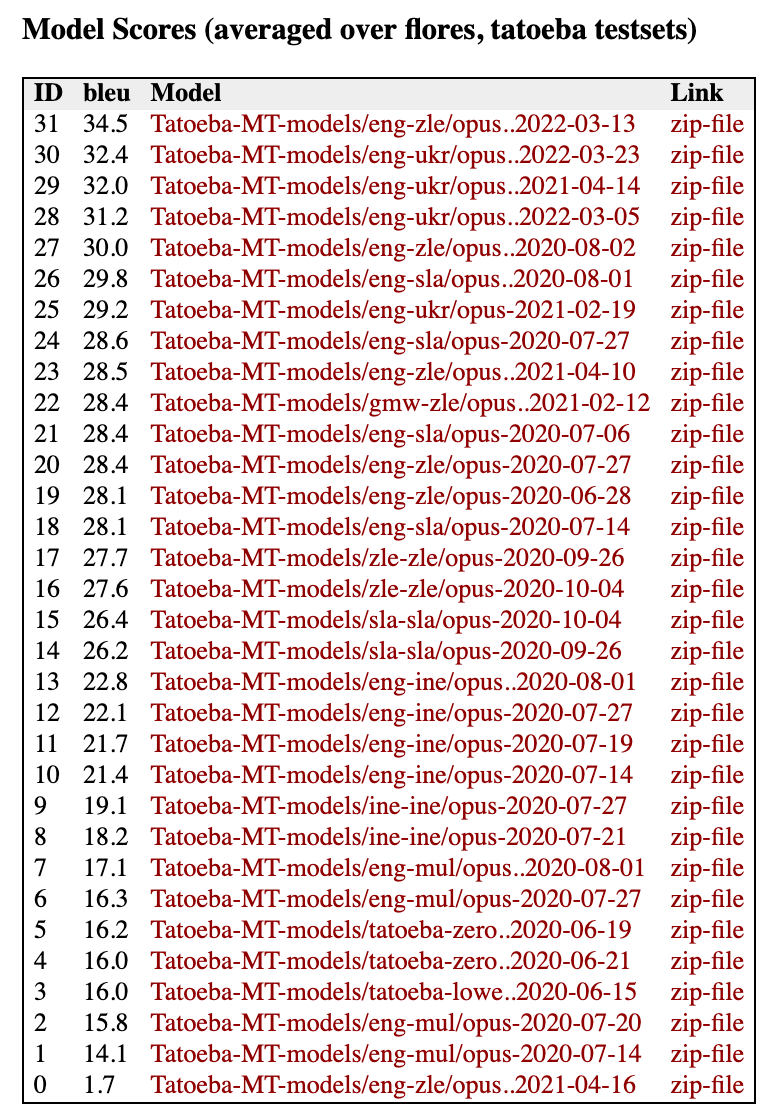}
    \caption{A screenshot from the OPUS-MT dashboard. The example shows averaged BLEU scores over two benchmarks (Flores and Tatoeba) for the translation from English to Ukrainian. Green bars mark compact student models that can be used for efficient translation.}
    \label{fig:opusmt-leaderboard}
\end{figure}

The large number of models we train and the high language coverage and diversity we support makes it necessary to monitor progress and to obtain an overview of the available systems that are among the released packages. A common way to summarize and compare models is to use leaderboards on established benchmarks. OPUS-MT implements a simple interactive dashboard that provides information from our regular benchmarks in terms of tables and bar charts. Figure~\ref{fig:opusmt-leaderboard} shows a screenshot from our website.\footnote{\url{https://opus.nlpl.eu/dashboard/}} with averaged BLEU scores for English--Ukrainian machine translation measured on Flores~\cite{goyal-etal-2022-flores} and the Tatoeba benchmark. Currently, we support BLEU~\citep{papineni-etal-2002-bleu}, spBLEU~\citep{goyal-etal-2022-flores}, chrF~\citep{popovic-2015-chrf}, chrf++~\citep{popovic-2017-chrf} and COMET~\citep{stewart-etal-2020-comet} scores but other measures may be added once they become available from our systematic test procedures. The dashboard allows to select language pairs, benchmarks and provides various views on language-specific or model-specific evaluations. The test set translations and the models themselves can be downloaded from the links provided in the table. The texts of the original benchmarks are also linked.

The dashboard takes the information from our score repository (OPUS-MT leaderboard) and currently does not support any dynamic upload of new models or translation results. The ambition is not to provide a fully-fledged system for benchmarking new systems but rather to provide a view on our evaluation results to provide summaries and overviews of OPUS-MT capabilities. Note that we include a large variety of benchmarks in order to provide a comprehensive picture on model performance, which is not restricted to one specific domain and evaluation test set.

Another useful visualization is the interactive geographic map that we generate from our released models.\footnote{\url{https://opus.nlpl.eu/NMT-map/}} Figure~\ref{fig:opusmt-map} in Section~\ref{sec:tatoeba} shows an example of such a plot. As we are striving for a wide language coverage, it is useful to visually see gaps across the globe. We, therefore, plot translation models according to their source or target language onto the geo-location provided by Glottolog.\footnote{\url{https://glottolog.org/}} We use the langinfo library\footnote{\url{https://github.com/robertostling/langinfo}} to extract the location information from the Glottolog database using the ISO-standard language IDs in our translation models. OpenStreetMap is used to visualize the locations on a map and we indicate the size of the test set by the size of the dot (to illustrate reliability of the score) and use colors on a continuous scale from green (best) to red (worst) to indicate quality in terms of the selected benchmark. 
We base the visualization on chrF scores, which is more reliable across languages than BLEU, but note that even those scores are problematic to compare in general.
Only the highest scoring model is shown for each language pair.

The map is interactive and allows to select source or target language to base the illustration on. We also generate maps for various benchmarks and use feature templates to show information about the model that correspond to a dot on the map. From that template, links to the downloadable model releases and further information are available.

\subsection{Linguistic Test Suites}
\label{sec:test-suites}

The impressive advances in translation quality seen in recent years have led to a discussion whether translations produced by professional human translators can still be distinguished from the output of NMT systems, and to what extent automatic evaluation measures can reliably account for these differences~\cite{hassan-human-parity,laubli-emnlp18,toral-wmt18}.
One answer to this question lies in the development of so-called \textit{test suites}~\cite{burchardt-pbml17} or \textit{challenge sets}~\cite{isabelle-emnlp17} that focus on particular linguistic phenomena that are known to be difficult to evaluate with simple reference-based metrics such as BLEU. However, most existing test suites require significant amounts of expert knowledge and manual work for compiling the examples, which typically limits their coverage to a small number of translation directions.

\begin{table}
\centering \small
\caption{Examples of English--German test suite instances. The ambiguous English source word is highlighted in bold, and correct and incorrect German translations -- as inferred by the MuCoW procedure -- are given.}
\label{tab:example-wmt}
\begin{tabular}{p{0.48\linewidth}p{0.2\linewidth}p{0.2\linewidth}}
\toprule
Example containing \textbf{ambiguous word} & Correct \newline translations & Incorrect \newline translations \\
\midrule
It occurred to me that my \textbf{watch} might be broken. & Armband\-uhr, Uhr & Wache \\
I hope you didn't get distracted during your \textbf{watch}. & Wache & Armband\-uhr, Uhr \\
\cmidrule(lr){1-3}
In winter, the dry leaves fly around in the \textbf{air}. & Luft, Luft\-raum, Aura & Miene, Ausdruck \\
He remained silent for a moment, with a thoughtful but contented \textbf{air}. & Miene, Ausdruck & Luft, Luft\-raum, Aura \\
\cmidrule(lr){1-3}
Harry had to back out of the competition because of a broken \textbf{arm}. & Arm & Waffe \\
So does the cop who left his side \textbf{arm} in a subway bathroom. & Waffe & Arm \\
\cmidrule(lr){1-3}
Drain the pasta and return the pasta to the \textbf{pot}. & Blumentopf, Kochtopf, Topf, Nachttopf & Marihuana, Gras \\
Where did those idiots get all of this \textbf{pot} anyhow? & Marihuana, Gras & Blumentopf, Kochtopf, Topf, Nachttopf \\
\bottomrule
\end{tabular}
\end{table}

To facilitate the development of test suites for a wide range of language pairs, we have introduced \textsc{MuCoW},\footnote{\url{https://github.com/Helsinki-NLP/MuCoW}} a language-independent method for automatically building test suites for lexically ambiguous nouns~\cite{mucow-wmt19,mucow-lrec,mucow-wmt20}. \textsc{MuCoW} takes advantage of the parallel corpora available in OPUS and of BabelNet~\cite{NavigliPonzetto:12aij}, a wide-coverage multilingual encyclopedic dictionary obtained automatically from various resources (WordNet and Wikipedia, among others). In a nutshell, the three following steps are needed to create a \textsc{MuCoW} test suite:
\begin{enumerate}
\item We identify ambiguous source words and their translations in parallel corpora, matching only those words that are found to be ambiguous in BabelNet.
\item Since lexical resources such as BabelNet are known to suffer from overly fine granularity of their sense inventory, we merge the BabelNet sense clusters with similar meanings, taking advantage of pre-trained sense embeddings: if the cosine similarity between two sense embeddings exceeds a certain threshold, the corresponding clusters are merged \citep{mucow-wmt19}. 
\item To build the test suite properly speaking, we extract sentence pairs from the parallel corpora, making sure that sentences from different corpora are represented. Each sentence is complemented with a list of correct and a list of incorrect translations of the ambiguous source word.
\end{enumerate}

Table~\ref{tab:example-wmt} shows some examples of test suite instances for the English--German translation direction. \textsc{MuCoW} has identified \textit{watch, air, arm} and \textit{pot} as ambiguous English nouns, extracted example sentences using these nouns from various OPUS corpora, and associated each sentence with correct and incorrect German translations.

MuCoW has been first introduced at WMT-19 as a test suite for the News Translation Task, for 9 language pairs~\cite{mucow-wmt19}. In this context, we showed that tuned NMT systems performed well on our evaluation suite, but struggled when dealing with out-of-domain data. We observed the same trend in the following year at WMT-20, with only a general improvement in translation quality for the top-ranked systems~\cite{mucow-wmt20}.

Finally, we also created a MuCoW benchmark set that includes training data with known sense distributions, to evaluate competing systems on a fair ground~\cite{mucow-lrec}. Our findings show that state-of-the-art Transformer models struggle when dealing with rare word senses. Interestingly enough, adding more training data, not necessarily containing the ambiguous words of interest, contributes to mitigating such issues. Moreover, we also show that word segmentation does not affect the disambiguation ability much, whereas the performance drops consistently across languages when evaluating sentences from noisy sources.

\section{Scaling-Up and Scaling-Down}
\label{sec:scaling}

Increasing language coverage and providing lightweight models that are easy to integrate and deploy are both our priorities for advancing the MT field towards a more open and inclusive state. However, pursuing both objectives in parallel without losing translation performance is still an open problem. 

A common approach to multilingual NMT makes use of fully shared models, where a single neural model is trained over parallel data including several translation directions, and all model weights are updated at every training step.
The multilingual models obtained with OPUS-MT as described in Section~\ref{sec:training-pipelines} are examples of this paradigm. 

Adding languages to a fully shared multilingual model~\citep{johnson-etal-2017-googles} implies distributing the model capacity over several translation tasks, which may lead to decreased bilingual performance if the number of parameters of the model is kept constant.
On the other side, having a large number of bilingual models, or a model architecture with language-specific components, can be impractical without extensive access to HPC facilities.
In the first case, the model size can become unmanageable due to the additional language-specific components.
In the second case, scaling up language coverage involves handling a large number of (relatively small) NMT model files. 
In this section, we discuss two strategies that mitigate this issue and give an overview of ongoing experiments.

\subsection{Modular NMT}
The OPUS ecosystem presents
extensive potential for exploring and analyzing the capabilities of multilingual MT systems.
Hand in hand with the European Research Council (ERC) funded \textit{Found in Translation} (FoTran) project,\footnote{\url{http://www.helsinki.fi/fotran}} we develop natural language understanding models trained on implicit information given by large collections of human translations.
Aligning both initiatives, it is in our best interest to distribute and make broadly available, both our toolkit for distributed training of multilingual models and the pre-trained models to be reused and also fine-tuned to new tasks. At training time, highly multilingual models require a large amount of computational power. However, we  are building an architecture with a modular design that allows to reuse the trained components (language-specific and shared modules) on relatively small processing units,\footnote{\url{https://github.com/Helsinki-NLP/FoTraNMT}} making multilingual models more affordable and increasing their applicability.
In contrast to the OPUS-MT models, this implementation is based on OpenNMT-py\footnote{\url{https://github.com/OpenNMT/OpenNMT-py}}~\citep{klein-etal-2020-opennmt}. Our ambition is also to release models from those efforts in the near future but for now, we focus on the development of the modular and scalable framework first.

\subsubsection{The Multilingual Attention-Bridge Model}
Our original motivation to build a multilingual translation model in FoTran is to explore the role of ``cross-lingual grounding'' for resolving ambiguities through translation. The intuition behind this idea is that translations provide a semantic mirror \citep{Dyvik2004TranslationsAS} reflecting the same meaning with the expressions of another language. We want to explore that signal in representation learning but it also becomes interesting in transfer learning for the original task of machine translation.
We designed a modular NN architecture~\cite{vazquez-etal-2019-multilingual} in which the model incorporates an intermediate shared layer that exploits the semantics from each language while keeping language-specific components.

\begin{figure}[tb]
    \centering
    \includegraphics[width=\textwidth]{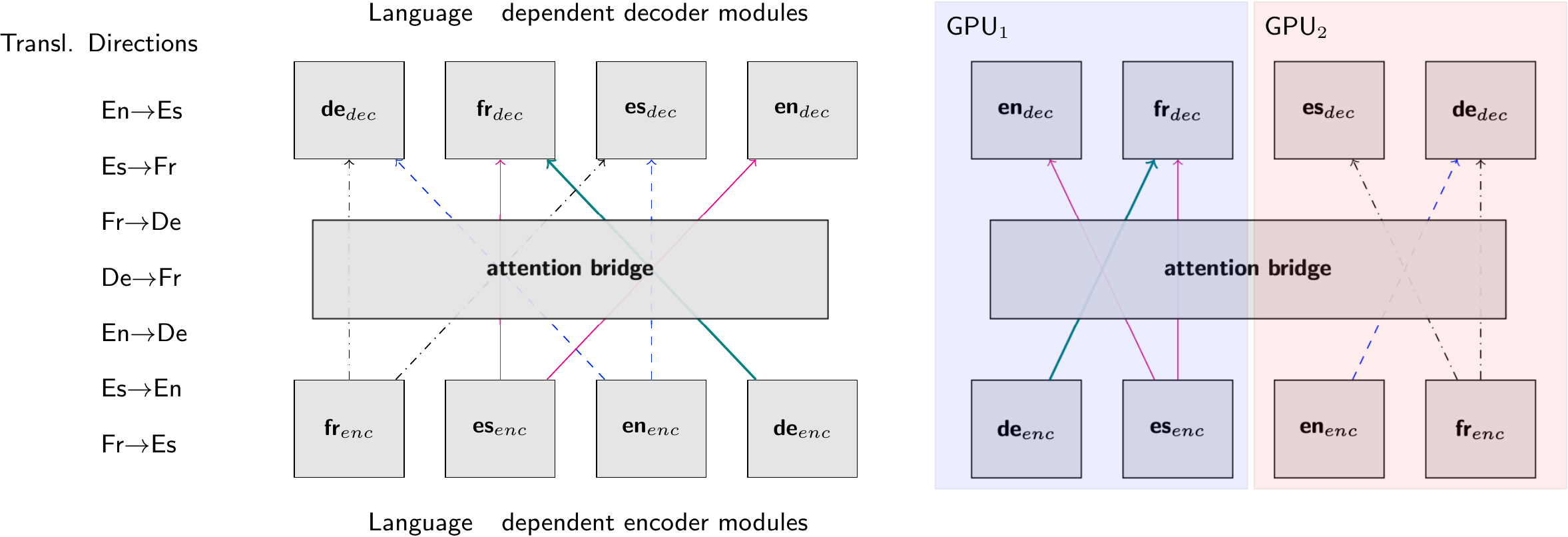}
    \caption{Diagram of the multilingual attention-bridge model used in a simple example. In this example we use 7 language pairs and 2 GPUs to illustrate the effectiveness of assigning language pairs in different GPUs to reduce inter-device communication. }\label{fig:ab_diagram}
\end{figure}

The architecture is illustrated in the central diagram of Figure~\ref{fig:ab_diagram}. It follows a traditional sequence-to-sequence encoder--decoder architecture, but incorporates language specific encoders and decoders that are connected through a shared component to enable multilingual training and knowledge transfer. We obtain multilingual representations generated by the encoders by forcing the information to flow through a bottleneck inner-attention layer connecting all the language-specific modules~\cite{vazquez-etal-2020-systematic}.
This layer summarizes the encoder information in a fixed-size (language-agnostic) meaning representation, which is useful for machine translation (including the support of zero-shot scenarios) and downstream tasks that require semantic reasoning and inference. 
Experimental results point towards the improvement of both the translation quality, and the abstractions acquired by our model when including more languages~\cite{vazquez-etal-2019-multilingual,raganato-etal-2019-evaluation}.

\subsubsection{Scaling Up for High Linguistic Diversity}
In its basic implementation, our multilingual model requires high computational resources at training time due to the use of language-specific modules. Scaling up the number of translation directions on a single device is restricted by the memory limits of that specific computing node.

To address those challenges, we implemented a multi-node and multi-GPU training setup that incorporates the following strategies:  (1)~distribute efficiently the modules across several processing units, (2)~train the network over many translation directions reducing memory overhead, and (3)~reuse the trained modules without having to load the entire network.
Taken together, these strategies deliver a cost-effective multilingual NMT system that can further be used for extracting multilingual meaning representations.

We distribute the model across multiple processing units by loading, in each device, encoders and decoders for only a subset of the translation directions.
The inner-attention layer is shared across all processing units.
All modules that are present in more than one device are initialized with the same weights. Parameters that are present in more than one device need to remain synchronous at each training step. We schedule the gradient communication of all parameters to reduce the inter communication load.

In general, allocating language pairs with common source/target languages on the same device decreases both the total memory footprint of the model and the amount of communication needed to keep the modules synced. To see this, we can follow the example in Figure~\ref{fig:ab_diagram}. In the example, there are 7 language pairs to be trained simultaneously, and we have access to 2 GPUs. Defining a partition like the one on the right-most minimizes the amount of communication between the devices to keep the modules synced, reducing the training time. 

Gathering together language pairs based on the source (or target) language could result in a scattered configuration depending on the target (or source) languages included. When dealing with a high number of translation directions (and a limited number of source and target languages) it becomes impossible to avoid this condition.
We address these problems using two strategies. First, we solve the allocation problem to minimize inter-device communication. Since in most cases the problem has no feasible exact solution, we approximate it using the Hungarian algorithm \cite{kuhn-1955-hungarian} over a cost matrix that makes it cheaper to assign the same language to a given GPU. 
Second, we propose to schedule the gradient updates to minimize the waiting time when inter-device communication happens. 

With this infrastructure at hand, we currently work on scaling up experiments to highly multilingual models that we can train on large high-performance clusters with a wide distribution of components over compute nodes and massively parallel training data. We aim at a large language coverage and will also release the model with separate components that can individually be loaded for efficient inference and further fine-tuning or downstream testing.

\subsection{Knowledge Distillation}
\label{sec:opusmt-distill}

Machine translation requires optimization for speed and size to be available in practical solutions and on various devices. OPUS-MT models are already compact and fast in comparison to the ever-growing multilingual language models that are often referred to in recent NLP work. Nevertheless, there are various ways to further improve decoding speed while reducing resource requirements. Knowledge distillation \cite{hintonEA-2015} is one of the popular techniques that reduces complex neural architectures (used as a so-called teacher model) to compact student models, which obtain the essential information from the generalization the teacher model has learned previously in a typically very expensive and time-consuming learning process.

\begin{figure}[tb]
    \centering
    \includegraphics[width=\textwidth]{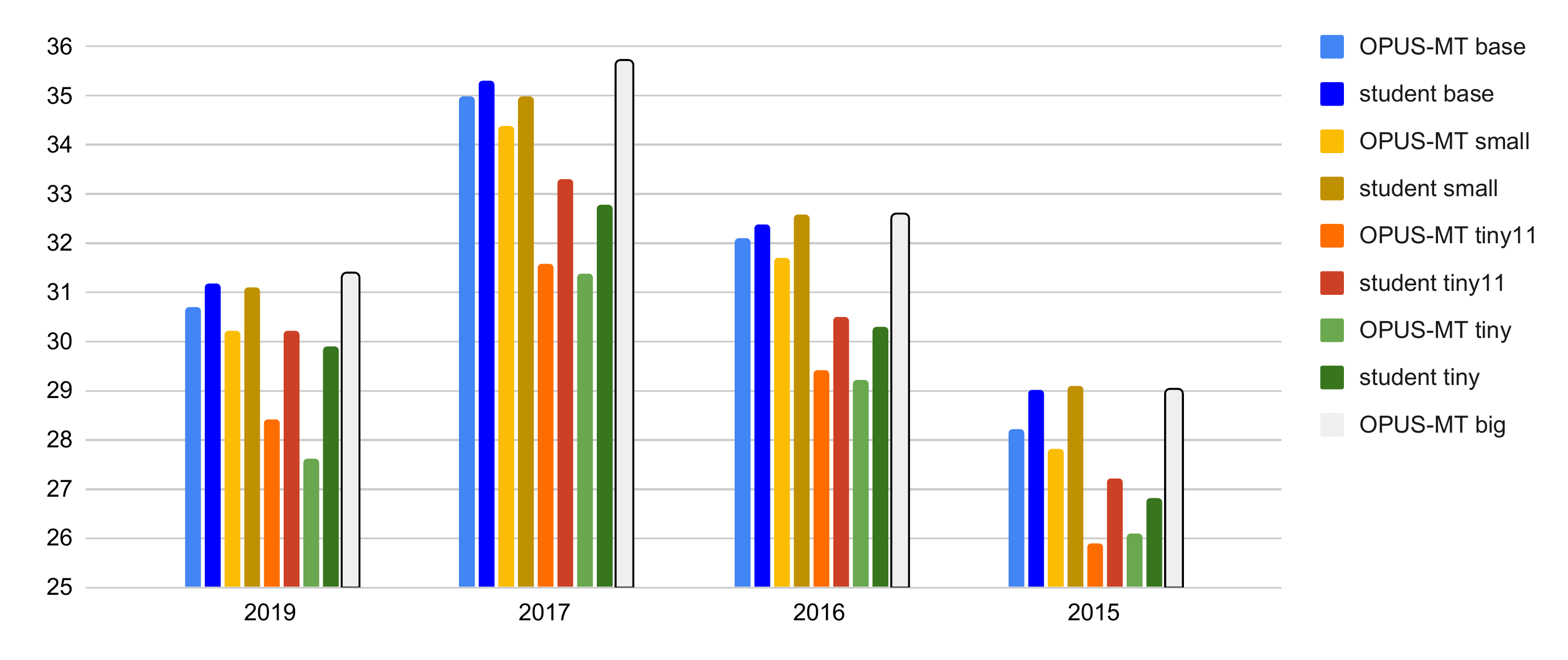}
    \caption{Comparing regular NMT models (labeled as OPUS-MT) to distilled student models created through sequence-level knowledge distillation for Finnish--English MT models of various sizes. All distilled models use the same teacher model (transformer-big, last bar in each group) and regular models are trained from scratch with the same architectures as the corresponding student models. The years in the X-axis refer to different news test sets from WMT and the scores refer to BLEU scores.}
    \label{fig:opusmt-distill}
\end{figure}

We use sequence-level knowledge distillation~\cite{kim-rush-2016-sequence}, in which a teacher model provides complete translations of some training data, and the student model learns the task on that output rather than the original reference translations. For simplicity, 
we ``forward-translate'' the training data with the teacher model using a narrow-width beam search and only use the best translation for teaching the student. 
Currently, we do not apply ensemble methods either, which would push the output quality of the teacher a bit further. However, we apply normalized cross-entropy filtering~\cite{behnke-etal-2021-efficient} to remove some translation noise using a reverse translation model to score the translations obtained by the teacher model. This score indicates the level of ``hallucination'', i.e.\ how much the translation diverges from the original input in a sense that it cannot reliably be reconstructed from the translation produced by the system. Following related work, we retain 95\% of the data that have been sorted by that score.

In our current experiments, we look at different network architectures to study the impact of distillation. In particular, we use a base transformer model with 6 layers in the decoder and then reduce the decoder from a 6-layer transformer model to an RNN-based variant using simpler simple recurrent units (SSRU)~\cite{Kim-2019} with two stacked layers (henceforth called ``small"). Following previous work~\cite{behnke-etal-2021-efficient}, we also test the reduction of the embedding size to 256 dimensions (from 512 in default settings) and try two different variants (``tiny" and ``tiny11") that differ in the size of the transformer feed-forward network parameter (1,536 in ``tiny11" instead of the default 2,048 in ``tiny") and the size of the encoder (3 layers in ``tiny" instead of 6).

A drop in performance can be expected when reducing the network and the parameters size of the model. However, student models are known to recover well from the reduced capacity and, therefore, create efficient alternatives to more expensive models. Figure~\ref{fig:opusmt-distill} shows our results in terms of BLEU scores for Finnish--English MT for the various models we tried in comparison to standard non-distilled ones. We can see that in all cases the regular model of corresponding size is outperformed by the student distilled from a larger teacher model and, especially important, smaller student models substantially recover from the drop in performance we see with reduced model capacities in regular training from scratch. In particular, small students reach more or less the same performance as the much bigger teacher model, which is a remarkable achievement.

\begin{table}
\begin{center}
\begin{minipage}{220pt}
    \caption{Comparison of different models for translating from Finnish to English. Decoding speed is measured on the Tatoeba test set with 10,000 sentences covering approximately 48,000 words. Model names refer to the same settings as in Figure~\ref{fig:opusmt-distill}. Model size refers to gzipped file size (with and without model quantization). Decoding is done on a CPU node with 4 cores with and without lexical shortlists (see Section~\ref{sec:applications}).}
    \label{tab:opusmt-students}
\begin{tabular}{@{\extracolsep{\fill}}ccccc@{\extracolsep{\fill}}}
\toprule
model   & \multicolumn{2}{c}{compressed model size} & \multicolumn{2}{c}{decoding speed}\\ \cmidrule(lr){2-3} \cmidrule(lr){4-5}%
& original  & quantized & 4 CPUs & +shortlist \\
\midrule
big     & 891 MB    & 224 MB &          & \\
base    & 294 MB    & 74 MB & 46.36s    & 40.47s\\
small   & 226 MB    & 57 MB & 24.07s    & 17.89s\\
tiny11  & 96 MB     & 25 MB & 10.98s    & 7.24s\\
tiny    & 89 MB     & 23 MB & 9.90s     & 6.22s\\
\botrule
\end{tabular}
\end{minipage}
\end{center}
\end{table}

The effect of model reduction can be seen in terms of size and decoding speed. Table~\ref{tab:opusmt-students} summarizes properties and benchmarks on the Tatoeba test set with 10,000 sentences and over 48,000 words for the Finnish--English models discussed above. The space requirements go down dramatically and the smallest model is just about 23 MB in size, about 10\% of the big transformer model we used as a teacher (12 layers each in encoder and decoder). We also see the importance of quantization (using 8-bit integers, \texttt{int8}, in this case). Another substantial improvement can be seen in decoding speed: The smallest model is able to decode the entire test set in less than 10 seconds on a 4-core CPU machine, an increase in speed by more than factor 4 compared to the base transformer model. Additional lexical shortlists, compiled from 100 top aligned tokens on the training data, make it possible to push the time down to about 6 seconds.

With these levels of performance and size, we can afford real-time translation on regular and even small devices. The browser-based MT solutions and local desktop apps discussed in Section~\ref{sec:opusmt-integration} become feasible solutions with such highly-optimized distilled student models and the models are compatible with the Bergamot-derived systems and can immediately applied in applications based on that project (see, e.g.\ \url{https://translatelocally.com/web/}).

\section{Related Work}
\label{sec:relatedwork}

OPUS and OPUS-MT are certainly not alone on the quest for open and transparent machine translation. Numerous projects and initiatives have been created over the years and listing them all is beyond the scope of this section. One of the issues with project-related work is the long-term perspective and the creation of sustainable platforms with a clear continuation.

During the era of statistical machine translation, many projects have been related to the framework of Moses,\footnote{\url{http://www2.statmt.org/moses/}} leading to a rich infrastructure of resources and tools. Projects to be mentioned include EuroMatrix, EuroMatrixPlus,\footnote{\url{https://www.euromatrixplus.net/}} TC-STAR,\footnote{\url{http://www.tcstar.org/}} LetsMT!\footnote{\url{http://project.letsmt.eu/}} and MateCAT.\footnote{\url{https://www.matecat.com/}}

Another long-lasting infrastructure from the paradigm of rule-based machine translation is connected to Apertium.\footnote{\url{https://www.apertium.org/}} Compared to most developments driven by EU projects that typically result in commercial exploitation, the development of Apertium always emphasizes its place as an open-source and free translation platform. In that sense, it is the most related work to OPUS-MT, but with a focus on a different translation approach and a much longer history.

OPUS and OPUS-MT build on top of many other open-source and data-sharing efforts. EU-funded research projects and resource development supported the creation of resources and tools in projects like ParaCrawl,\footnote{\url{https://www.paracrawl.eu/}} GoURMET\footnote{\url{https://gourmet-project.eu/}}, MT4ALL\footnote{\url{http://ixa2.si.ehu.eus/mt4all/project.html}} and MaCoCu.\footnote{\url{https://macocu.eu/}}

Open software platforms are also essential for our project and the rapid development of neural MT could not have been possible without software packages such as OpenNMT\footnote{\url{https://opennmt.net/}} and Marian-NMT,\footnote{\url{https://marian-nmt.github.io/}} to name just two of them with relevance to our project. Translation efficiency was a focus in the Bergamot project,\footnote{\url{https://browser.mt/}} and the work continues in the HPLT project,\footnote{\url{https://hplt-project.org/}} which has close ties to OPUS and OPUS-MT.

Language coverage in NLP and inclusivity are also addressed by various regional and international initiatives. Masakhane,\footnote{\url{https://www.masakhane.io/}} for example, is a grass-roots organization for African NLP, which includes dedicated work on resource and tool development for translating African languages. From the Nordic perspective, one can mention Giellatekno,\footnote{\url{https://giellatekno.uit.no}} with their focus on Sami language technology. The EU also tries to push for more comprehensive data sharing and resource building with the European Language Resource Coordination (ELRC)\footnote{\url{https://lr-coordination.eu/}} and their digital strategy on creating accessible data spaces. Resource repositories like Meta-Share\footnote{\url{http://www.meta-share.org/}} have been developed and filled with information, links and metadata, and resource infrastructures such as CLARIN\footnote{\url{https://www.clarin.eu/}} play an important role in the coordination of such efforts.

Another push for open machine translation certainly also comes from commercial research labs and NLP startups. Research at big tech companies supports the development of platforms such as fairseq,\footnote{\url{https://github.com/facebookresearch/fairseq}} tensor2tensor\footnote{\url{https://github.com/tensorflow/tensor2tensor}} and Sockeye.\footnote{\url{https://github.com/awslabs/sockeye}} Important multilingual resources such as WikiMatrix and CCMatrix have their origin in those labs and made their way into OPUS. The No Language Left Behind project\footnote{\url{https://ai.facebook.com/research/no-language-left-behind/}} together with extended benchmarks\footnote{\url{https://github.com/facebookresearch/flores}} further pushed the frontiers of multilingual machine translation.  Another central point for NLP research nowadays is Hugging Face together with their popular \texttt{transformers} library and the growing collection of data sets\footnote{\url{https://huggingface.co/datasets}} and models.\footnote{\url{https://huggingface.co/models}}
The availability of all resources in one hub makes it easy and straightforward to get started with NLP research.

Another recent initiative to be mentioned in connection with machine translation is LibreTranslate.\footnote{\url{https://libretranslate.com/}} Their efforts are very much in line with OPUS and OPUS-MT, and we will explore collaboration possibilities in future work to join their and other great open-source initiatives that appear in modern NLP.

\section{Conclusions}

This paper provides an overview of OPUS-MT and its embedding into the OPUS ecosystem. We describe various components that facilitate data curation, model development and machine translation integration into various platforms and applications. Our initiative emphasizes a large language coverage and focuses on public resources and open-source solutions in order to create a transparent and widely applicable support for machine translation. The efforts of OPUS-MT already produced a large number of high-quality pre-trained NMT models that are ready to be used and adapted to various needs in research and practical application development. With this, the project supports a sustainable infrastructure that enables reuse of computationally expensive components.
Giving access to the entire output of our project enables us to make efficient machine translation available for a wide range of users and NLP developers, without the need of high-performance IT infrastructure to train complex neural models from scratch. Democratizing MT in this way is a major step in the direction of an inclusive information society where language barriers do not lead to significant disadvantages. OPUS-MT contributes to this mission with its 
community-driven open source initiative.

Our future plans include the development of practical solutions for modular NMT in a highly multilingual setup and further advances in transfer learning and model efficiency. We also continue our efforts in data collection and curation and aim to further increase language support and coverage. Furthermore, we also want to include paragraph- or document-level translation models in OPUS-MT and integrate other advances that can be pushed into production-ready solutions.

\section*{Acknowledgements}
The research presented in this paper was supported by the FoTran project, funded by the European Research Council (ERC) under the European Union’s Horizon 2020 research and innovation program (grant agreement no. 771113),
the European Language Grid project through its open call for pilot projects with funding from the European Union’s Horizon 2020 Research and Innovation program under grant agreement no. 825627 (ELG) and
the Swedish Culture Foundation (Svenska Kulturfonden) in Finland under the grant agreement no. 139592. 
Continued support is provided by the EU Horizon project HPLT (grant agreement no. 101070350) and GreenNLP funded by the Academy of Finland (project ID 353164).
The work was also supported by the NVIDIA AI Technology Center (NVAITC) and we would like to thank NVIDIA for their hardware grants providing GPU cards for research and development.
Finally, we like to acknowledge the great support by CSC, the Finnish IT Center for Science Ltd., providing extensive computing and storage facilities used in this research and all projects associated with it.

\bibliography{opus}


\begin{thebibliography}{56}
\ifx \bisbn   \undefined \def \bisbn  #1{ISBN #1}\fi
\ifx \binits  \undefined \def \binits#1{#1}\fi
\ifx \bauthor  \undefined \def \bauthor#1{#1}\fi
\ifx \batitle  \undefined \def \batitle#1{#1}\fi
\ifx \bjtitle  \undefined \def \bjtitle#1{#1}\fi
\ifx \bvolume  \undefined \def \bvolume#1{\textbf{#1}}\fi
\ifx \byear  \undefined \def \byear#1{#1}\fi
\ifx \bissue  \undefined \def \bissue#1{#1}\fi
\ifx \bfpage  \undefined \def \bfpage#1{#1}\fi
\ifx \blpage  \undefined \def \blpage #1{#1}\fi
\ifx \burl  \undefined \def \burl#1{\textsf{#1}}\fi
\ifx \doiurl  \undefined \def \doiurl#1{\url{https://doi.org/#1}}\fi
\ifx \betal  \undefined \def \betal{\textit{et al.}}\fi
\ifx \binstitute  \undefined \def \binstitute#1{#1}\fi
\ifx \binstitutionaled  \undefined \def \binstitutionaled#1{#1}\fi
\ifx \bctitle  \undefined \def \bctitle#1{#1}\fi
\ifx \beditor  \undefined \def \beditor#1{#1}\fi
\ifx \bpublisher  \undefined \def \bpublisher#1{#1}\fi
\ifx \bbtitle  \undefined \def \bbtitle#1{#1}\fi
\ifx \bedition  \undefined \def \bedition#1{#1}\fi
\ifx \bseriesno  \undefined \def \bseriesno#1{#1}\fi
\ifx \blocation  \undefined \def \blocation#1{#1}\fi
\ifx \bsertitle  \undefined \def \bsertitle#1{#1}\fi
\ifx \bsnm \undefined \def \bsnm#1{#1}\fi
\ifx \bsuffix \undefined \def \bsuffix#1{#1}\fi
\ifx \bparticle \undefined \def \bparticle#1{#1}\fi
\ifx \barticle \undefined \def \barticle#1{#1}\fi
\bibcommenthead
\ifx \bconfdate \undefined \def \bconfdate #1{#1}\fi
\ifx \botherref \undefined \def \botherref #1{#1}\fi
\ifx \url \undefined \def \url#1{\textsf{#1}}\fi
\ifx \bchapter \undefined \def \bchapter#1{#1}\fi
\ifx \bbook \undefined \def \bbook#1{#1}\fi
\ifx \bcomment \undefined \def \bcomment#1{#1}\fi
\ifx \oauthor \undefined \def \oauthor#1{#1}\fi
\ifx \citeauthoryear \undefined \def \citeauthoryear#1{#1}\fi
\ifx \endbibitem  \undefined \def \endbibitem {}\fi
\ifx \bconflocation  \undefined \def \bconflocation#1{#1}\fi
\ifx \arxivurl  \undefined \def \arxivurl#1{\textsf{#1}}\fi
\csname PreBibitemsHook\endcsname

\bibitem{tiedemann-nygaard-2004-opus}
\begin{bchapter}
\bauthor{\bsnm{Tiedemann}, \binits{J.}},
\bauthor{\bsnm{Nygaard}, \binits{L.}}:
\bctitle{The {OPUS} corpus - parallel and free:
  \url{http://logos.uio.no/opus}}.
In: \bbtitle{Proceedings of the Fourth International Conference on Language
  Resources and Evaluation ({LREC}{'}04)},
pp. \bfpage{1183}--\blpage{1186}.
\bpublisher{European Language Resources Association (ELRA)},
\blocation{Lisbon, Portugal}
(\byear{2004}).
\burl{\url{http://www.lrec-conf.org/proceedings/lrec2004/pdf/320.pdf}}
\end{bchapter}
\endbibitem

\bibitem{tiedemann-thottingal-2020-opus}
\begin{bchapter}
\bauthor{\bsnm{Tiedemann}, \binits{J.}},
\bauthor{\bsnm{Thottingal}, \binits{S.}}:
\bctitle{{OPUS}-{MT} {--} building open translation services for the world}.
In: \bbtitle{Proceedings of the 22nd Annual Conference of the European
  Association for Machine Translation},
pp. \bfpage{479}--\blpage{480}.
\bpublisher{European Association for Machine Translation},
\blocation{Lisboa, Portugal}
(\byear{2020}).
\burl{\url{https://aclanthology.org/2020.eamt-1.61}}
\end{bchapter}
\endbibitem

\bibitem{tiedemann-2009-news}
\begin{barticle}
\bauthor{\bsnm{Tiedemann}, \binits{J.}}:
\batitle{News from {OPUS} - a collection of multilingual parallel corpora with
  tools and interfaces}.
\bjtitle{Recent Advances in Natural Language Processing}
\bvolume{V},
\bfpage{237}--\blpage{248}
(\byear{2009})
\end{barticle}
\endbibitem

\bibitem{tiedemann-2012-parallel}
\begin{bchapter}
\bauthor{\bsnm{Tiedemann}, \binits{J.}}:
\bctitle{Parallel data, tools and interfaces in {OPUS}}.
In: \bbtitle{Proceedings of the Eighth International Conference on Language
  Resources and Evaluation ({LREC}'12)},
pp. \bfpage{2214}--\blpage{2218}.
\bpublisher{European Language Resources Association (ELRA)},
\blocation{Istanbul, Turkey}
(\byear{2012}).
\burl{\url{http://www.lrec-conf.org/proceedings/lrec2012/pdf/463\_Paper.pdf}}
\end{bchapter}
\endbibitem

\bibitem{zipf}
\begin{bbook}
\bauthor{\bsnm{Zipf}, \binits{G.K.}}:
\bbtitle{Selected Studies of the Principle of Relative Frequency in Language}.
\bpublisher{Harvard University Press},
\blocation{Cambridge, MA and London, England}
(\byear{1932}).
\burl{\url{https://doi.org/10.4159/harvard.9780674434929}}
\end{bbook}
\endbibitem

\bibitem{koehn-etal-2007-moses}
\begin{bchapter}
\bauthor{\bsnm{Koehn}, \binits{P.}},
\bauthor{\bsnm{Hoang}, \binits{H.}},
\bauthor{\bsnm{Birch}, \binits{A.}},
\bauthor{\bsnm{Callison-Burch}, \binits{C.}},
\bauthor{\bsnm{Federico}, \binits{M.}},
\bauthor{\bsnm{Bertoldi}, \binits{N.}},
\bauthor{\bsnm{Cowan}, \binits{B.}},
\bauthor{\bsnm{Shen}, \binits{W.}},
\bauthor{\bsnm{Moran}, \binits{C.}},
\bauthor{\bsnm{Zens}, \binits{R.}},
\bauthor{\bsnm{Dyer}, \binits{C.}},
\bauthor{\bsnm{Bojar}, \binits{O.}},
\bauthor{\bsnm{Constantin}, \binits{A.}},
\bauthor{\bsnm{Herbst}, \binits{E.}}:
\bctitle{{M}oses: Open source toolkit for statistical machine translation}.
In: \bbtitle{Proceedings of the 45th Annual Meeting of the Association for
  Computational Linguistics Companion Volume Proceedings of the Demo and Poster
  Sessions},
pp. \bfpage{177}--\blpage{180}.
\bpublisher{Association for Computational Linguistics},
\blocation{Prague, Czech Republic}
(\byear{2007}).
\burl{\url{https://aclanthology.org/P07-2045}}
\end{bchapter}
\endbibitem

\bibitem{aulamo-etal-2020-opustools}
\begin{bchapter}
\bauthor{\bsnm{Aulamo}, \binits{M.}},
\bauthor{\bsnm{Sulubacak}, \binits{U.}},
\bauthor{\bsnm{Virpioja}, \binits{S.}},
\bauthor{\bsnm{Tiedemann}, \binits{J.}}:
\bctitle{{O}pus{T}ools and parallel corpus diagnostics}.
In: \bbtitle{Proceedings of the 12th Language Resources and Evaluation
  Conference},
pp. \bfpage{3782}--\blpage{3789}.
\bpublisher{European Language Resources Association},
\blocation{Marseille, France}
(\byear{2020}).
\burl{\url{https://aclanthology.org/2020.lrec-1.467}}
\end{bchapter}
\endbibitem

\bibitem{lui-baldwin-2012-langid}
\begin{bchapter}
\bauthor{\bsnm{Lui}, \binits{M.}},
\bauthor{\bsnm{Baldwin}, \binits{T.}}:
\bctitle{langid.py: An off-the-shelf language identification tool}.
In: \bbtitle{Proceedings of the {ACL} 2012 System Demonstrations},
pp. \bfpage{25}--\blpage{30}.
\bpublisher{Association for Computational Linguistics},
\blocation{Jeju Island, Korea}
(\byear{2012}).
\burl{\url{https://aclanthology.org/P12-3005}}
\end{bchapter}
\endbibitem

\bibitem{aulamo-etal-2020-opusfilter}
\begin{bchapter}
\bauthor{\bsnm{Aulamo}, \binits{M.}},
\bauthor{\bsnm{Virpioja}, \binits{S.}},
\bauthor{\bsnm{Tiedemann}, \binits{J.}}:
\bctitle{{O}pus{F}ilter: A configurable parallel corpus filtering toolbox}.
In: \bbtitle{Proceedings of the 58th Annual Meeting of the Association for
  Computational Linguistics: System Demonstrations},
pp. \bfpage{150}--\blpage{156}.
\bpublisher{Association for Computational Linguistics},
\blocation{Online}
(\byear{2020}).
\burl{\url{https://aclanthology.org/2020.acl-demos.20}}
\end{bchapter}
\endbibitem

\bibitem{sanchez-cartagena-etal-2018-prompsit}
\begin{bchapter}
\bauthor{\bsnm{S{\'a}nchez-Cartagena}, \binits{V.M.}},
\bauthor{\bsnm{Ba{\~n}{\'o}n}, \binits{M.}},
\bauthor{\bsnm{Ortiz-Rojas}, \binits{S.}},
\bauthor{\bsnm{Ram{\'\i}rez}, \binits{G.}}:
\bctitle{Prompsit{'}s submission to {WMT} 2018 parallel corpus filtering shared
  task}.
In: \bbtitle{Proceedings of the Third Conference on Machine Translation: Shared
  Task Papers},
pp. \bfpage{955}--\blpage{962}.
\bpublisher{Association for Computational Linguistics},
\blocation{Belgium, Brussels}
(\byear{2018}).
\burl{\url{https://aclanthology.org/W18-6488}}
\end{bchapter}
\endbibitem

\bibitem{xu-koehn-2017-zipporah}
\begin{bchapter}
\bauthor{\bsnm{Xu}, \binits{H.}},
\bauthor{\bsnm{Koehn}, \binits{P.}}:
\bctitle{{Z}ipporah: a fast and scalable data cleaning system for noisy
  web-crawled parallel corpora}.
In: \bbtitle{Proceedings of the 2017 Conference on Empirical Methods in Natural
  Language Processing},
pp. \bfpage{2945}--\blpage{2950}.
\bpublisher{Association for Computational Linguistics},
\blocation{Copenhagen, Denmark}
(\byear{2017}).
\burl{\url{https://www.aclweb.org/anthology/D17-1319}}
\end{bchapter}
\endbibitem

\bibitem{marie-etal-2021-scientific}
\begin{bchapter}
\bauthor{\bsnm{Marie}, \binits{B.}},
\bauthor{\bsnm{Fujita}, \binits{A.}},
\bauthor{\bsnm{Rubino}, \binits{R.}}:
\bctitle{Scientific credibility of machine translation research: A
  meta-evaluation of 769 papers}.
In: \bbtitle{Proceedings of the 59th Annual Meeting of the Association for
  Computational Linguistics and the 11th International Joint Conference on
  Natural Language Processing (Volume 1: Long Papers)},
pp. \bfpage{7297}--\blpage{7306}.
\bpublisher{Association for Computational Linguistics},
\blocation{Online}
(\byear{2021}).
\burl{\url{https://aclanthology.org/2021.acl-long.566}}
\end{bchapter}
\endbibitem

\bibitem{sennrich-etal-2016-neural}
\begin{bchapter}
\bauthor{\bsnm{Sennrich}, \binits{R.}},
\bauthor{\bsnm{Haddow}, \binits{B.}},
\bauthor{\bsnm{Birch}, \binits{A.}}:
\bctitle{Neural machine translation of rare words with subword units}.
In: \bbtitle{Proceedings of the 54th Annual Meeting of the Association for
  Computational Linguistics (Volume 1: Long Papers)},
pp. \bfpage{1715}--\blpage{1725}.
\bpublisher{Association for Computational Linguistics},
\blocation{Berlin, Germany}
(\byear{2016}).
\burl{\url{https://aclanthology.org/P16-1162}}
\end{bchapter}
\endbibitem

\bibitem{virpioja-etal-2013-morfessor}
\begin{botherref}
\oauthor{\bsnm{Virpioja}, \binits{S.}},
\oauthor{\bsnm{Smit}, \binits{P.}},
\oauthor{\bsnm{Gr\"{o}nroos}, \binits{S.-A.}},
\oauthor{\bsnm{Kurimo}, \binits{M.}}:
Morfessor 2.0: Python implementation and extensions for {M}orfessor {B}aseline.
Report 25/2013 in Aalto University publication series SCIENCE + TECHNOLOGY,
Department of Signal Processing and Acoustics, Aalto University,
Helsinki, Finland
(2013)
\end{botherref}
\endbibitem

\bibitem{joulin-etal-2016-fasttext}
\begin{botherref}
\oauthor{\bsnm{Joulin}, \binits{A.}},
\oauthor{\bsnm{Grave}, \binits{E.}},
\oauthor{\bsnm{Bojanowski}, \binits{P.}},
\oauthor{\bsnm{Douze}, \binits{M.}},
\oauthor{\bsnm{Jégou}, \binits{H.}},
\oauthor{\bsnm{Mikolov}, \binits{T.}}:
FastText.zip: Compressing text classification models.
arXiv
(2016).
\url{https://arxiv.org/abs/1612.03651}
\end{botherref}
\endbibitem

\bibitem{joulin-etal-2017-bag}
\begin{bchapter}
\bauthor{\bsnm{Joulin}, \binits{A.}},
\bauthor{\bsnm{Grave}, \binits{E.}},
\bauthor{\bsnm{Bojanowski}, \binits{P.}},
\bauthor{\bsnm{Mikolov}, \binits{T.}}:
\bctitle{Bag of tricks for efficient text classification}.
In: \bbtitle{Proceedings of the 15th Conference of the {E}uropean Chapter of
  the Association for Computational Linguistics: Volume 2, Short Papers},
pp. \bfpage{427}--\blpage{431}.
\bpublisher{Association for Computational Linguistics},
\blocation{Valencia, Spain}
(\byear{2017}).
\burl{\url{https://aclanthology.org/E17-2068}}
\end{bchapter}
\endbibitem

\bibitem{siivola-etal-2007-growing}
\begin{barticle}
\bauthor{\bsnm{Siivola}, \binits{V.}},
\bauthor{\bsnm{Hirsim\"aki}, \binits{T.}},
\bauthor{\bsnm{Virpioja}, \binits{S.}}:
\batitle{On growing and pruning {K}neser-{N}ey smoothed n-gram models}.
\bjtitle{IEEE Transactions on Audio, Speech and Language Processing}
\bvolume{15}(\bissue{5}),
\bfpage{1617}--\blpage{1624}
(\byear{2007})
\end{barticle}
\endbibitem

\bibitem{ostling-tiedemann-2016-efficient}
\begin{barticle}
\bauthor{\bsnm{{\"O}stling}, \binits{R.}},
\bauthor{\bsnm{Tiedemann}, \binits{J.}}:
\batitle{Efficient word alignment with {M}arkov {C}hain {M}onte {C}arlo}.
\bjtitle{The Prague Bulletin of Mathematical Linguistics}
\bvolume{106},
\bfpage{125}--\blpage{146}
(\byear{2016})
\end{barticle}
\endbibitem

\bibitem{artetxe-schwenk-2019-margin}
\begin{bchapter}
\bauthor{\bsnm{Artetxe}, \binits{M.}},
\bauthor{\bsnm{Schwenk}, \binits{H.}}:
\bctitle{Margin-based parallel corpus mining with multilingual sentence
  embeddings}.
In: \bbtitle{Proceedings of the 57th Annual Meeting of the Association for
  Computational Linguistics},
pp. \bfpage{3197}--\blpage{3203}.
\bpublisher{Association for Computational Linguistics},
\blocation{Florence, Italy}
(\byear{2019}).
\burl{\url{https://aclanthology.org/P19-1309}}
\end{bchapter}
\endbibitem

\bibitem{vazquez-etal-2019-university}
\begin{bchapter}
\bauthor{\bsnm{V{\'a}zquez}, \binits{R.}},
\bauthor{\bsnm{Sulubacak}, \binits{U.}},
\bauthor{\bsnm{Tiedemann}, \binits{J.}}:
\bctitle{The {U}niversity of {H}elsinki submission to the {WMT}19 parallel
  corpus filtering task}.
In: \bbtitle{Proceedings of the Fourth Conference on Machine Translation
  (Volume 3: Shared Task Papers, Day 2)},
pp. \bfpage{294}--\blpage{300}.
\bpublisher{Association for Computational Linguistics},
\blocation{Florence, Italy}
(\byear{2019}).
\burl{\url{https://aclanthology.org/W19-5441}}
\end{bchapter}
\endbibitem

\bibitem{tiedemann-2020-tatoeba}
\begin{bchapter}
\bauthor{\bsnm{Tiedemann}, \binits{J.}}:
\bctitle{The {T}atoeba translation challenge {--} realistic data sets for low
  resource and multilingual {MT}}.
In: \bbtitle{Proceedings of the Fifth Conference on Machine Translation},
pp. \bfpage{1174}--\blpage{1182}.
\bpublisher{Association for Computational Linguistics},
\blocation{Online}
(\byear{2020}).
\burl{\url{https://aclanthology.org/2020.wmt-1.139}}
\end{bchapter}
\endbibitem

\bibitem{junczys-dowmunt-etal-2018-marian}
\begin{bchapter}
\bauthor{\bsnm{Junczys-Dowmunt}, \binits{M.}},
\bauthor{\bsnm{Grundkiewicz}, \binits{R.}},
\bauthor{\bsnm{Dwojak}, \binits{T.}},
\bauthor{\bsnm{Hoang}, \binits{H.}},
\bauthor{\bsnm{Heafield}, \binits{K.}},
\bauthor{\bsnm{Neckermann}, \binits{T.}},
\bauthor{\bsnm{Seide}, \binits{F.}},
\bauthor{\bsnm{Germann}, \binits{U.}},
\bauthor{\bsnm{Fikri~Aji}, \binits{A.}},
\bauthor{\bsnm{Bogoychev}, \binits{N.}},
\bauthor{\bsnm{Martins}, \binits{A.F.T.}},
\bauthor{\bsnm{Birch}, \binits{A.}}:
\bctitle{Marian: Fast neural machine translation in {C++}}.
In: \bbtitle{Proceedings of ACL 2018, System Demonstrations},
pp. \bfpage{116}--\blpage{121}.
\bpublisher{Association for Computational Linguistics},
\blocation{Melbourne, Australia}
(\byear{2018}).
\burl{\url{http://www.aclweb.org/anthology/P18-4020}}
\end{bchapter}
\endbibitem

\bibitem{kudo-richardson-2018-sentencepiece}
\begin{bchapter}
\bauthor{\bsnm{Kudo}, \binits{T.}},
\bauthor{\bsnm{Richardson}, \binits{J.}}:
\bctitle{{S}entence{P}iece: A simple and language independent subword tokenizer
  and detokenizer for neural text processing}.
In: \bbtitle{Proceedings of the 2018 Conference on Empirical Methods in Natural
  Language Processing: System Demonstrations},
pp. \bfpage{66}--\blpage{71}.
\bpublisher{Association for Computational Linguistics},
\blocation{Brussels, Belgium}
(\byear{2018}).
\burl{\url{https://aclanthology.org/D18-2012}}
\end{bchapter}
\endbibitem

\bibitem{dou2021word}
\begin{bchapter}
\bauthor{\bsnm{Dou}, \binits{Z.-Y.}},
\bauthor{\bsnm{Neubig}, \binits{G.}}:
\bctitle{Word alignment by fine-tuning embeddings on parallel corpora}.
In: \bbtitle{Proceedings of the 16th Conference of the European Chapter of the
  Association for Computational Linguistics: Main Volume},
pp. \bfpage{2112}--\blpage{2128}.
\bpublisher{Association for Computational Linguistics},
\blocation{Online}
(\byear{2021}).
\burl{\url{https://aclanthology.org/2021.eacl-main.181}}
\end{bchapter}
\endbibitem

\bibitem{arxiv.1907.05019}
\begin{botherref}
\oauthor{\bsnm{Arivazhagan}, \binits{N.}},
\oauthor{\bsnm{Bapna}, \binits{A.}},
\oauthor{\bsnm{Firat}, \binits{O.}},
\oauthor{\bsnm{Lepikhin}, \binits{D.}},
\oauthor{\bsnm{Johnson}, \binits{M.}},
\oauthor{\bsnm{Krikun}, \binits{M.}},
\oauthor{\bsnm{Chen}, \binits{M.X.}},
\oauthor{\bsnm{Cao}, \binits{Y.}},
\oauthor{\bsnm{Foster}, \binits{G.}},
\oauthor{\bsnm{Cherry}, \binits{C.}},
\oauthor{\bsnm{Macherey}, \binits{W.}},
\oauthor{\bsnm{Chen}, \binits{Z.}},
\oauthor{\bsnm{Wu}, \binits{Y.}}:
Massively Multilingual Neural Machine Translation in the Wild: Findings and
  Challenges.
arXiv
(2019).
\url{https://arxiv.org/abs/1907.05019}
\end{botherref}
\endbibitem

\bibitem{sennrich-etal-2016-improving}
\begin{bchapter}
\bauthor{\bsnm{Sennrich}, \binits{R.}},
\bauthor{\bsnm{Haddow}, \binits{B.}},
\bauthor{\bsnm{Birch}, \binits{A.}}:
\bctitle{Improving neural machine translation models with monolingual data}.
In: \bbtitle{Proceedings of the 54th Annual Meeting of the Association for
  Computational Linguistics (Volume 1: Long Papers)},
pp. \bfpage{86}--\blpage{96}.
\bpublisher{Association for Computational Linguistics},
\blocation{Berlin, Germany}
(\byear{2016}).
\burl{\url{https://aclanthology.org/P16-1009}}
\end{bchapter}
\endbibitem

\bibitem{hoang-etal-2018-iterative}
\begin{bchapter}
\bauthor{\bsnm{Hoang}, \binits{V.C.D.}},
\bauthor{\bsnm{Koehn}, \binits{P.}},
\bauthor{\bsnm{Haffari}, \binits{G.}},
\bauthor{\bsnm{Cohn}, \binits{T.}}:
\bctitle{Iterative back-translation for neural machine translation}.
In: \bbtitle{Proceedings of the 2nd Workshop on Neural Machine Translation and
  Generation},
pp. \bfpage{18}--\blpage{24}.
\bpublisher{Association for Computational Linguistics},
\blocation{Melbourne, Australia}
(\byear{2018}).
\burl{\url{https://aclanthology.org/W18-2703}}
\end{bchapter}
\endbibitem

\bibitem{kudo-2018-subword}
\begin{bchapter}
\bauthor{\bsnm{Kudo}, \binits{T.}}:
\bctitle{Subword regularization: Improving neural network translation models
  with multiple subword candidates}.
In: \bbtitle{Proceedings of the 56th Annual Meeting of the Association for
  Computational Linguistics (Volume 1: Long Papers)},
pp. \bfpage{66}--\blpage{75}.
\bpublisher{Association for Computational Linguistics},
\blocation{Melbourne, Australia}
(\byear{2018}).
\burl{\url{https://aclanthology.org/P18-1007}}
\end{bchapter}
\endbibitem

\bibitem{laubli-etal-2019-post}
\begin{bchapter}
\bauthor{\bsnm{L{\"a}ubli}, \binits{S.}},
\bauthor{\bsnm{Amrhein}, \binits{C.}},
\bauthor{\bsnm{D{\"u}ggelin}, \binits{P.}},
\bauthor{\bsnm{Gonzalez}, \binits{B.}},
\bauthor{\bsnm{Zwahlen}, \binits{A.}},
\bauthor{\bsnm{Volk}, \binits{M.}}:
\bctitle{Post-editing productivity with neural machine translation: An
  empirical assessment of speed and quality in the banking and finance domain}.
In: \bbtitle{Proceedings of Machine Translation Summit XVII: Research Track},
pp. \bfpage{267}--\blpage{272}.
\bpublisher{European Association for Machine Translation},
\blocation{Dublin, Ireland}
(\byear{2019}).
\burl{\url{https://aclanthology.org/W19-6626}}
\end{bchapter}
\endbibitem

\bibitem{informatics7020012}
\begin{botherref}
\oauthor{\bsnm{Macken}, \binits{L.}},
\oauthor{\bsnm{Prou}, \binits{D.}},
\oauthor{\bsnm{Tezcan}, \binits{A.}}:
Quantifying the effect of machine translation in a high-quality human
  translation production process.
Informatics
\textbf{7}(2)
(2020).
\doiurl{10.3390/informatics7020012}
\end{botherref}
\endbibitem

\bibitem{bergmanis-pinnis-2021-facilitating}
\begin{bchapter}
\bauthor{\bsnm{Bergmanis}, \binits{T.}},
\bauthor{\bsnm{Pinnis}, \binits{M.}}:
\bctitle{Facilitating terminology translation with target lemma annotations}.
In: \bbtitle{Proceedings of the 16th Conference of the European Chapter of the
  Association for Computational Linguistics: Main Volume},
pp. \bfpage{3105}--\blpage{3111}.
\bpublisher{Association for Computational Linguistics},
\blocation{Online}
(\byear{2021}).
\burl{\url{https://aclanthology.org/2021.eacl-main.271}}
\end{bchapter}
\endbibitem

\bibitem{goyal-etal-2022-flores}
\begin{barticle}
\bauthor{\bsnm{Goyal}, \binits{N.}},
\bauthor{\bsnm{Gao}, \binits{C.}},
\bauthor{\bsnm{Chaudhary}, \binits{V.}},
\bauthor{\bsnm{Chen}, \binits{P.-J.}},
\bauthor{\bsnm{Wenzek}, \binits{G.}},
\bauthor{\bsnm{Ju}, \binits{D.}},
\bauthor{\bsnm{Krishnan}, \binits{S.}},
\bauthor{\bsnm{Ranzato}, \binits{M.}},
\bauthor{\bsnm{Guzm{\'a}n}, \binits{F.}},
\bauthor{\bsnm{Fan}, \binits{A.}}:
\batitle{The {F}lores-101 evaluation benchmark for low-resource and
  multilingual machine translation}.
\bjtitle{Transactions of the Association for Computational Linguistics}
\bvolume{10},
\bfpage{522}--\blpage{538}
(\byear{2022}).
\doiurl{10.1162/tacl_a_00474}
\end{barticle}
\endbibitem

\bibitem{papineni-etal-2002-bleu}
\begin{bchapter}
\bauthor{\bsnm{Papineni}, \binits{K.}},
\bauthor{\bsnm{Roukos}, \binits{S.}},
\bauthor{\bsnm{Ward}, \binits{T.}},
\bauthor{\bsnm{Zhu}, \binits{W.-J.}}:
\bctitle{{B}leu: a method for automatic evaluation of machine translation}.
In: \bbtitle{Proceedings of the 40th Annual Meeting of the Association for
  Computational Linguistics},
pp. \bfpage{311}--\blpage{318}.
\bpublisher{Association for Computational Linguistics},
\blocation{Philadelphia, Pennsylvania, USA}
(\byear{2002}).
\burl{\url{https://aclanthology.org/P02-1040}}
\end{bchapter}
\endbibitem

\bibitem{popovic-2015-chrf}
\begin{bchapter}
\bauthor{\bsnm{Popovi{\'c}}, \binits{M.}}:
\bctitle{chr{F}: character n-gram {F}-score for automatic {MT} evaluation}.
In: \bbtitle{Proceedings of the Tenth Workshop on Statistical Machine
  Translation},
pp. \bfpage{392}--\blpage{395}.
\bpublisher{Association for Computational Linguistics},
\blocation{Lisbon, Portugal}
(\byear{2015}).
\burl{\url{https://aclanthology.org/W15-3049}}
\end{bchapter}
\endbibitem

\bibitem{popovic-2017-chrf}
\begin{bchapter}
\bauthor{\bsnm{Popovi{\'c}}, \binits{M.}}:
\bctitle{chr{F}++: words helping character n-grams}.
In: \bbtitle{Proceedings of the Second Conference on Machine Translation},
pp. \bfpage{612}--\blpage{618}.
\bpublisher{Association for Computational Linguistics},
\blocation{Copenhagen, Denmark}
(\byear{2017}).
\burl{\url{https://aclanthology.org/W17-4770}}
\end{bchapter}
\endbibitem

\bibitem{stewart-etal-2020-comet}
\begin{bchapter}
\bauthor{\bsnm{Stewart}, \binits{C.}},
\bauthor{\bsnm{Rei}, \binits{R.}},
\bauthor{\bsnm{Farinha}, \binits{C.}},
\bauthor{\bsnm{Lavie}, \binits{A.}}:
\bctitle{{COMET} - deploying a new state-of-the-art {MT} evaluation metric in
  production}.
In: \bbtitle{Proceedings of the 14th Conference of the Association for Machine
  Translation in the Americas (Volume 2: User Track)},
pp. \bfpage{78}--\blpage{109}.
\bpublisher{Association for Machine Translation in the Americas},
\blocation{Virtual}
(\byear{2020}).
\burl{\url{https://aclanthology.org/2020.amta-user.4}}
\end{bchapter}
\endbibitem

\bibitem{hassan-human-parity}
\begin{botherref}
\oauthor{\bsnm{Hassan}, \binits{H.}},
\oauthor{\bsnm{Aue}, \binits{A.}},
\oauthor{\bsnm{Chen}, \binits{C.}},
\oauthor{\bsnm{Chowdhary}, \binits{V.}},
\oauthor{\bsnm{Clark}, \binits{J.}},
\oauthor{\bsnm{Federmann}, \binits{C.}},
\oauthor{\bsnm{Huang}, \binits{X.}},
\oauthor{\bsnm{Junczys-Dowmunt}, \binits{M.}},
\oauthor{\bsnm{Lewis}, \binits{W.}},
\oauthor{\bsnm{Li}, \binits{M.}},
\oauthor{\bsnm{Liu}, \binits{S.}},
\oauthor{\bsnm{Liu}, \binits{T.-Y.}},
\oauthor{\bsnm{Luo}, \binits{R.}},
\oauthor{\bsnm{Menezes}, \binits{A.}},
\oauthor{\bsnm{Qin}, \binits{T.}},
\oauthor{\bsnm{Seide}, \binits{F.}},
\oauthor{\bsnm{Tan}, \binits{X.}},
\oauthor{\bsnm{Tian}, \binits{F.}},
\oauthor{\bsnm{Wu}, \binits{L.}},
\oauthor{\bsnm{Wu}, \binits{S.}},
\oauthor{\bsnm{Xia}, \binits{Y.}},
\oauthor{\bsnm{Zhang}, \binits{D.}},
\oauthor{\bsnm{Zhang}, \binits{Z.}},
\oauthor{\bsnm{Zhou}, \binits{M.}}:
Achieving Human Parity on Automatic Chinese to English News Translation.
arXiv
(2018).
\url{https://arxiv.org/abs/1803.05567}
\end{botherref}
\endbibitem

\bibitem{laubli-emnlp18}
\begin{bchapter}
\bauthor{\bsnm{L{\"a}ubli}, \binits{S.}},
\bauthor{\bsnm{Sennrich}, \binits{R.}},
\bauthor{\bsnm{Volk}, \binits{M.}}:
\bctitle{Has machine translation achieved human parity? a case for
  document-level evaluation}.
In: \bbtitle{Proceedings of the 2018 Conference on Empirical Methods in Natural
  Language Processing},
pp. \bfpage{4791}--\blpage{4796}.
\bpublisher{Association for Computational Linguistics},
\blocation{Brussels, Belgium}
(\byear{2018}).
\burl{\url{https://aclanthology.org/D18-1512}}
\end{bchapter}
\endbibitem

\bibitem{toral-wmt18}
\begin{bchapter}
\bauthor{\bsnm{Toral}, \binits{A.}},
\bauthor{\bsnm{Castilho}, \binits{S.}},
\bauthor{\bsnm{Hu}, \binits{K.}},
\bauthor{\bsnm{Way}, \binits{A.}}:
\bctitle{Attaining the unattainable? reassessing claims of human parity in
  neural machine translation}.
In: \bbtitle{Proceedings of the Third Conference on Machine Translation:
  Research Papers},
pp. \bfpage{113}--\blpage{123}.
\bpublisher{Association for Computational Linguistics},
\blocation{Brussels, Belgium}
(\byear{2018}).
\burl{\url{https://aclanthology.org/W18-6312}}
\end{bchapter}
\endbibitem

\bibitem{burchardt-pbml17}
\begin{barticle}
\bauthor{\bsnm{Burchardt}, \binits{A.}},
\bauthor{\bsnm{Macketanz}, \binits{V.}},
\bauthor{\bsnm{Dehdari}, \binits{J.}},
\bauthor{\bsnm{Heigold}, \binits{G.}},
\bauthor{\bsnm{Peter}, \binits{J.-T.}},
\bauthor{\bsnm{Williams}, \binits{P.}}:
\batitle{A linguistic evaluation of rule-based, phrase-based, and neural {MT}
  engines}.
\bjtitle{The Prague Bulletin of Mathematical Linguistics}
\bvolume{108},
\bfpage{159}--\blpage{170}
(\byear{2017})
\end{barticle}
\endbibitem

\bibitem{isabelle-emnlp17}
\begin{bchapter}
\bauthor{\bsnm{Isabelle}, \binits{P.}},
\bauthor{\bsnm{Cherry}, \binits{C.}},
\bauthor{\bsnm{Foster}, \binits{G.}}:
\bctitle{A challenge set approach to evaluating machine translation}.
In: \bbtitle{Proceedings of the 2017 Conference on Empirical Methods in Natural
  Language Processing},
pp. \bfpage{2486}--\blpage{2496}.
\bpublisher{Association for Computational Linguistics},
\blocation{Copenhagen, Denmark}
(\byear{2017}).
\burl{\url{https://aclanthology.org/D17-1263}}
\end{bchapter}
\endbibitem

\bibitem{mucow-wmt19}
\begin{bchapter}
\bauthor{\bsnm{Raganato}, \binits{A.}},
\bauthor{\bsnm{Scherrer}, \binits{Y.}},
\bauthor{\bsnm{Tiedemann}, \binits{J.}}:
\bctitle{The {M}u{C}o{W} test suite at {WMT} 2019: Automatically harvested
  multilingual contrastive word sense disambiguation test sets for machine
  translation}.
In: \bbtitle{Proceedings of the Fourth Conference on Machine Translation
  (Volume 2: Shared Task Papers, Day 1)},
pp. \bfpage{470}--\blpage{480}.
\bpublisher{Association for Computational Linguistics},
\blocation{Florence, Italy}
(\byear{2019}).
\burl{\url{https://www.aclweb.org/anthology/W19-5354}}
\end{bchapter}
\endbibitem

\bibitem{mucow-lrec}
\begin{bchapter}
\bauthor{\bsnm{Raganato}, \binits{A.}},
\bauthor{\bsnm{Scherrer}, \binits{Y.}},
\bauthor{\bsnm{Tiedemann}, \binits{J.}}:
\bctitle{An evaluation benchmark for testing the word sense disambiguation
  capabilities of machine translation systems}.
In: \bbtitle{Proceedings of The 12th Language Resources and Evaluation
  Conference},
pp. \bfpage{3668}--\blpage{3675}.
\bpublisher{European Language Resources Association},
\blocation{Marseille, France}
(\byear{2020}).
\burl{\url{https://www.aclweb.org/anthology/2020.lrec-1.452}}
\end{bchapter}
\endbibitem

\bibitem{mucow-wmt20}
\begin{bchapter}
\bauthor{\bsnm{Scherrer}, \binits{Y.}},
\bauthor{\bsnm{Raganato}, \binits{A.}},
\bauthor{\bsnm{Tiedemann}, \binits{J.}}:
\bctitle{The {MUCOW} word sense disambiguation test suite at {WMT} 2020}.
In: \bbtitle{Proceedings of the Fifth Conference on Machine Translation},
pp. \bfpage{365}--\blpage{370}.
\bpublisher{Association for Computational Linguistics},
\blocation{Online}
(\byear{2020}).
\burl{\url{https://aclanthology.org/2020.wmt-1.40}}
\end{bchapter}
\endbibitem

\bibitem{NavigliPonzetto:12aij}
\begin{barticle}
\bauthor{\bsnm{Navigli}, \binits{R.}},
\bauthor{\bsnm{Ponzetto}, \binits{S.P.}}:
\batitle{{B}abel{N}et: {T}he automatic construction, evaluation and application
  of a wide-coverage multilingual semantic network}.
\bjtitle{Artificial Intelligence}
\bvolume{193},
\bfpage{217}--\blpage{250}
(\byear{2012})
\end{barticle}
\endbibitem

\bibitem{johnson-etal-2017-googles}
\begin{barticle}
\bauthor{\bsnm{Johnson}, \binits{M.}},
\bauthor{\bsnm{Schuster}, \binits{M.}},
\bauthor{\bsnm{Le}, \binits{Q.V.}},
\bauthor{\bsnm{Krikun}, \binits{M.}},
\bauthor{\bsnm{Wu}, \binits{Y.}},
\bauthor{\bsnm{Chen}, \binits{Z.}},
\bauthor{\bsnm{Thorat}, \binits{N.}},
\bauthor{\bsnm{Vi{\'e}gas}, \binits{F.}},
\bauthor{\bsnm{Wattenberg}, \binits{M.}},
\bauthor{\bsnm{Corrado}, \binits{G.}},
\bauthor{\bsnm{Hughes}, \binits{M.}},
\bauthor{\bsnm{Dean}, \binits{J.}}:
\batitle{{G}oogle{'}s multilingual neural machine translation system: Enabling
  zero-shot translation}.
\bjtitle{Transactions of the Association for Computational Linguistics}
\bvolume{5},
\bfpage{339}--\blpage{351}
(\byear{2017}).
\doiurl{10.1162/tacl_a_00065}
\end{barticle}
\endbibitem

\bibitem{klein-etal-2020-opennmt}
\begin{bchapter}
\bauthor{\bsnm{Klein}, \binits{G.}},
\bauthor{\bsnm{Hernandez}, \binits{F.}},
\bauthor{\bsnm{Nguyen}, \binits{V.}},
\bauthor{\bsnm{Senellart}, \binits{J.}}:
\bctitle{The {O}pen{NMT} neural machine translation toolkit: 2020 edition}.
In: \bbtitle{Proceedings of the 14th Conference of the Association for Machine
  Translation in the Americas (Volume 1: Research Track)},
pp. \bfpage{102}--\blpage{109}.
\bpublisher{Association for Machine Translation in the Americas},
\blocation{Virtual}
(\byear{2020}).
\burl{\url{https://aclanthology.org/2020.amta-research.9}}
\end{bchapter}
\endbibitem

\bibitem{Dyvik2004TranslationsAS}
\begin{bchapter}
\bauthor{\bsnm{Dyvik}, \binits{H.}}:
\bctitle{Translations as semantic mirrors: from parallel corpus to wordnet}.
In: \bbtitle{Advances in Corpus Linguistics, Papers from the 23rd International
  Conference on English Language Research on Computerized Corpora (ICAME)},
vol. \bseriesno{49},
pp. \bfpage{309}--\blpage{326}.
\bpublisher{Brill},
\blocation{Gothenburg, Sweden}
(\byear{2004}).
\doiurl{10.1163/9789004333710_019}
\end{bchapter}
\endbibitem

\bibitem{vazquez-etal-2019-multilingual}
\begin{bchapter}
\bauthor{\bsnm{V{\'a}zquez}, \binits{R.}},
\bauthor{\bsnm{Raganato}, \binits{A.}},
\bauthor{\bsnm{Tiedemann}, \binits{J.}},
\bauthor{\bsnm{Creutz}, \binits{M.}}:
\bctitle{Multilingual {NMT} with a language-independent attention bridge}.
In: \bbtitle{Proceedings of the 4th Workshop on Representation Learning for NLP
  (RepL4NLP-2019)},
pp. \bfpage{33}--\blpage{39}.
\bpublisher{Association for Computational Linguistics},
\blocation{Florence, Italy}
(\byear{2019}).
\burl{\url{https://aclanthology.org/W19-4305}}
\end{bchapter}
\endbibitem

\bibitem{vazquez-etal-2020-systematic}
\begin{barticle}
\bauthor{\bsnm{V{\'a}zquez}, \binits{R.}},
\bauthor{\bsnm{Raganato}, \binits{A.}},
\bauthor{\bsnm{Creutz}, \binits{M.}},
\bauthor{\bsnm{Tiedemann}, \binits{J.}}:
\batitle{A systematic study of inner-attention-based sentence representations
  in multilingual neural machine translation}.
\bjtitle{Computational Linguistics}
\bvolume{46}(\bissue{2}),
\bfpage{387}--\blpage{424}
(\byear{2020}).
\doiurl{10.1162/coli_a_00377}
\end{barticle}
\endbibitem

\bibitem{raganato-etal-2019-evaluation}
\begin{bchapter}
\bauthor{\bsnm{Raganato}, \binits{A.}},
\bauthor{\bsnm{V{\'a}zquez}, \binits{R.}},
\bauthor{\bsnm{Creutz}, \binits{M.}},
\bauthor{\bsnm{Tiedemann}, \binits{J.}}:
\bctitle{An evaluation of language-agnostic inner-attention-based
  representations in machine translation}.
In: \bbtitle{Proceedings of the 4th Workshop on Representation Learning for NLP
  (RepL4NLP-2019)},
pp. \bfpage{27}--\blpage{32}.
\bpublisher{Association for Computational Linguistics},
\blocation{Florence, Italy}
(\byear{2019}).
\burl{\url{https://aclanthology.org/W19-4304}}
\end{bchapter}
\endbibitem

\bibitem{kuhn-1955-hungarian}
\begin{barticle}
\bauthor{\bsnm{Kuhn}, \binits{H.W.}}:
\batitle{The {H}ungarian method for the assignment problem}.
\bjtitle{Naval Research Logistics Quarterly}
\bvolume{2},
\bfpage{83}--\blpage{97}
(\byear{1955})
\end{barticle}
\endbibitem

\bibitem{hintonEA-2015}
\begin{botherref}
\oauthor{\bsnm{Hinton}, \binits{G.}},
\oauthor{\bsnm{Vinyals}, \binits{O.}},
\oauthor{\bsnm{Dean}, \binits{J.}}:
Distilling the Knowledge in a Neural Network.
arXiv
(2015).
\url{https://arxiv.org/abs/1503.02531}
\end{botherref}
\endbibitem

\bibitem{kim-rush-2016-sequence}
\begin{bchapter}
\bauthor{\bsnm{Kim}, \binits{Y.}},
\bauthor{\bsnm{Rush}, \binits{A.M.}}:
\bctitle{Sequence-level knowledge distillation}.
In: \bbtitle{Proceedings of the 2016 Conference on Empirical Methods in Natural
  Language Processing},
pp. \bfpage{1317}--\blpage{1327}.
\bpublisher{Association for Computational Linguistics},
\blocation{Austin, Texas}
(\byear{2016}).
\burl{\url{https://aclanthology.org/D16-1139}}
\end{bchapter}
\endbibitem

\bibitem{behnke-etal-2021-efficient}
\begin{bchapter}
\bauthor{\bsnm{Behnke}, \binits{M.}},
\bauthor{\bsnm{Bogoychev}, \binits{N.}},
\bauthor{\bsnm{Aji}, \binits{A.F.}},
\bauthor{\bsnm{Heafield}, \binits{K.}},
\bauthor{\bsnm{Nail}, \binits{G.}},
\bauthor{\bsnm{Zhu}, \binits{Q.}},
\bauthor{\bsnm{Tchistiakova}, \binits{S.}},
\bauthor{\bparticle{van~der} \bsnm{Linde}, \binits{J.}},
\bauthor{\bsnm{Chen}, \binits{P.}},
\bauthor{\bsnm{Kashyap}, \binits{S.}},
\bauthor{\bsnm{Grundkiewicz}, \binits{R.}}:
\bctitle{Efficient machine translation with model pruning and quantization}.
In: \bbtitle{Proceedings of the Sixth Conference on Machine Translation},
pp. \bfpage{775}--\blpage{780}.
\bpublisher{Association for Computational Linguistics},
\blocation{Online}
(\byear{2021}).
\burl{\url{https://aclanthology.org/2021.wmt-1.74}}
\end{bchapter}
\endbibitem

\bibitem{Kim-2019}
\begin{bchapter}
\bauthor{\bsnm{Kim}, \binits{Y.J.}},
\bauthor{\bsnm{Junczys-Dowmunt}, \binits{M.}},
\bauthor{\bsnm{Hassan}, \binits{H.}},
\bauthor{\bsnm{Fikri~Aji}, \binits{A.}},
\bauthor{\bsnm{Heafield}, \binits{K.}},
\bauthor{\bsnm{Grundkiewicz}, \binits{R.}},
\bauthor{\bsnm{Bogoychev}, \binits{N.}}:
\bctitle{From research to production and back: Ludicrously fast neural machine
  translation}.
In: \bbtitle{Proceedings of the 3rd Workshop on Neural Generation and
  Translation},
pp. \bfpage{280}--\blpage{288}.
\bpublisher{Association for Computational Linguistics},
\blocation{Hong Kong}
(\byear{2019}).
\burl{\url{https://aclanthology.org/D19-5632}}
\end{bchapter}
\endbibitem

\end{thebibliography}


\end{document}